\documentclass{article} %

\PassOptionsToPackage{table}{xcolor}

\usepackage[T1]{fontenc}
\usepackage{arxiv_style,times}

\usepackage{amsmath,amsfonts,bm}

\def\eqref#1{equation~\ref{#1}}

\def\1{\bm{1}}

\DeclareMathAlphabet{\mathsfit}{\encodingdefault}{\sfdefault}{m}{sl}
\SetMathAlphabet{\mathsfit}{bold}{\encodingdefault}{\sfdefault}{bx}{n}

\definecolor{darkGreen}{RGB}{62, 168, 70}

\usepackage{amsmath}
\usepackage{amssymb}
\usepackage{amsthm}
\usepackage{graphicx}
\usepackage{xcolor}
\usepackage{booktabs}
\usepackage{multirow}
\usepackage{url}
\usepackage{tabularx}
\usepackage{array}
\usepackage[pagebackref,breaklinks,colorlinks,allcolors=darkGreen]{hyperref}

\usepackage{makecell}

\usepackage{natbib}
\usepackage{wrapfig}

\def\methodname{Dr.~Free}

\title{\centering
\methodname: You Don't Need Difficulty Rewards\\
for Self-Evolving Search Agents\par}
\author{
  \parbox{0.95\textwidth}{
    \centering
    \normalfont\small
    Zhipeng Qian$^{1,2}$\quad
    Zihan Liang$^{2}$\quad
    Yufei Ma$^{2}$\quad
    Jie Ma$^{3}$\quad
    Ben Chen$^{2}$\quad 
    Huangyu Dai$^{2}$\quad 
    Lingtao Mao$^{2}$\quad  \\
    Xinyu Sun$^{2}$ \quad 
    Tong zhao$^{2}$\quad
    Xuxin Zhang$^{2}$\quad
    Qingpeng Cai$^{2}$\quad
    Peng Jiang$^{2}$\quad
    Qibin Hou$^{1,\dagger}$\\[5pt]
    $^{1}$VCIP, School of Computer Science, Nankai University\\
    $^{2}$Kuaishou Technology\quad
    $^{3}$Xiamen University\\[3pt]
    $^{\dagger}$Corresponding author
  }
}

\iclrfinalcopy

\hypersetup{
  pdftitle={Dr. Free: You Don't Need Difficulty Rewards for Self-Evolving Search Agents},
  pdfauthor={Zhipeng Qian, Zihan Liang, Yufei Ma, Jie Ma, Ben Chen, Huangyu Dai, Lingtao Mao, Xinyu Sun, Tong zhao, Xuxin Zhang, Qingpeng Cai, Peng Jiang, Qibin Hou}
}

\begin{document}

\maketitle

\begin{abstract}
A central limitation of current data-free self-evolution methods for training search agents is their reliance on difficulty-based proposer rewards. These methods reward a proposer for generating questions that challenge a co-evolving solver, using solver difficulty as a proxy for question quality.
Yet difficulty alone is insufficient to distinguish questions that require cross-passage evidence from those that are answerable via simpler shortcuts.
In addition, measuring difficulty demands repeated solver rollouts for every candidate question, leading to substantial computational costs.
In this paper, we introduce \methodname, the first self-evolving search framework that eliminates difficulty-based proposer rewards and directly optimizes for evidence necessity relative to shortcut contexts.
\methodname{} samples relational chains from a knowledge graph and pairs them with aligned passages, giving question generation an explicit multi-hop structure. A generated question receives a positive information-gain reward only when the likelihood of the target answer under the complete evidence passages exceeds the maximum likelihood under all evaluated shortcut contexts. Because this signal is computed from teacher-forced likelihoods, it removes the need for pass-rate estimation and reduces proposer training time by over $7\times$. Experiments on seven open-domain QA benchmarks show that \methodname{} outperforms prior data-free search agents and the supervised baseline, with large improvements on multi-hop QA benchmarks.

\end{abstract}

\section{Introduction}
\label{sec:introduction}

Training search agents typically requires curated question-answer pairs, incurring high data costs~\citep{jin2025searchr1trainingllmsreason,kwiatkowski2019natural,yang2018hotpotqa,trivedi2023interleaving,li2025search,qian2026plan}.%
Self-evolving methods reduce this dependence through proposer-solver self-play, in which a proposer continually generates questions, a solver learns to answer them, and solver performance in turn provides the training signal for subsequent question generation, enabling the two models to co-evolve over successive rounds~\citep{zhao2026absolute,huang2025r}.
Search Self-Play~\citep{lu2025search} generates multi-turn search questions sampled from public-QA datasets and filters the collected evidence with RAG, whereas Dr.~Zero~\citep{yue2026dr} removes curated QA pairs by generating questions from corpus passages and rewarding proposer difficulty. Subsequent methods incorporate knowledge graphs for question construction~\citep{wu2026knowledge,tan2026searchmaster} or reward the marginal utility of grounded evidence~\citep{arai2026eve}.
Despite these advances, these methods still rely on solver difficulty as the primary training signal, even though difficulty does not reveal whether a question genuinely requires the intended evidence.

The core issue is that solver difficulty and evidence necessity measure fundamentally different properties of a question-answer pair.
Difficulty is inherently solver-relative: It indicates whether the current model can produce a correct answer~\citep{yue2026dr}, rather than assessing the construction quality of the question itself.
This distinction is particularly important for multi-hop QA, where models can often reach the correct answer through reasoning shortcuts or disconnected evidence rather than the intended reasoning chain~\citep{jiang2019avoiding,trivedi2020multihop,guo2023counterfactual,li2026annotations}.
A well-constructed multi-hop question should instead require the integration of information across multiple passages, such that no individual passage alone provides sufficient evidence for the answer~\citep{ho2020constructing,trivedi2022musique,press2023measuring,feng2026group}.

\begin{figure*}[!t]
\centering
\setlength{\abovecaptionskip}{2pt}
\includegraphics[width=\columnwidth]{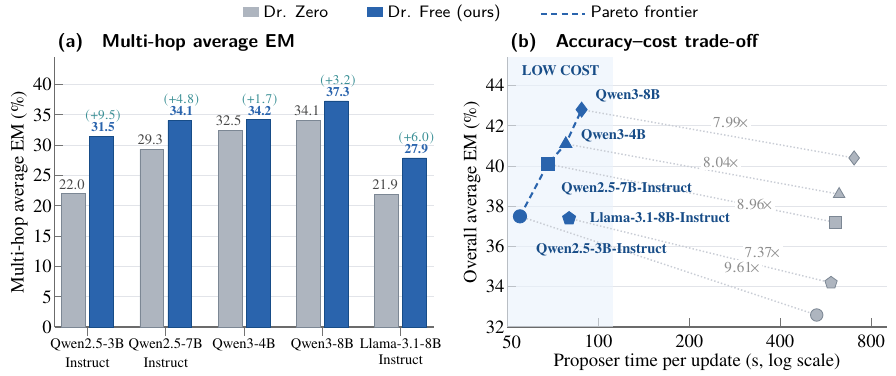}
\vspace{-0.5cm}
\caption{\textbf{\methodname{} outperforms Dr.~Zero across five backbones while training faster.}}
\label{fig:teaser}
\vspace{-0.6cm}
\end{figure*}

\begin{figure*}[!t]
\centering
\setlength{\abovecaptionskip}{2pt}
\includegraphics[width=\columnwidth]
{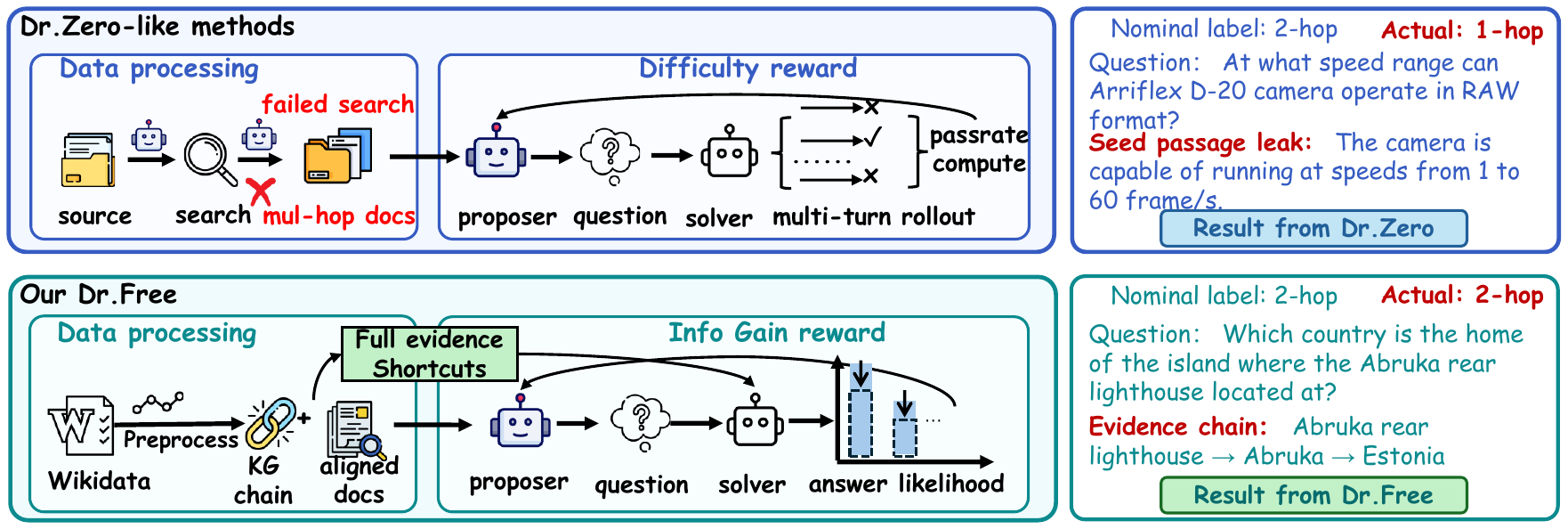}
\vspace{-0.3cm}
\caption{Comparison between Dr.~Zero-like methods and \methodname{}. The key differences lie in question construction and proposer reward design. Dr.~Zero-like methods construct questions through proposer-side search and optimize the proposer with rollout-based difficulty rewards, which may permit zero-search generation with target answers exposed in seed passages. In contrast, \methodname{} constructs questions from verified knowledge-graph chains with aligned passages and replaces difficulty-based rewards with a shortcut-aware information-gain objective that directly promotes dependence on cross-passage evidence.}
\label{fig:method_compar}
\end{figure*}

Our audit of Dr.~Zero reproductions further confirms this failure mode in practice. Across two backbone models, more than $80\%$ of questions labeled as multi-hop are generated without search, and the target answer for most questions already appears in the seed passage (Table~\ref{tab:diag}).
As shown in Figure~\ref{fig:method_compar}, a representative Dr.~Zero trace illustrates this mismatch. The question is labelled as two-hop, although the proposer performs no search and the target answer is directly available in the seed passage. Because the difficulty reward is computed solely from solver rollout outcomes, it cannot detect this evidence shortcut.
Consequently, hop labels often reflect the requested generation format rather than demonstrated cross-document evidence composition, leaving the resulting question-answer pairs susceptible to single-passage shortcuts.
Moreover, solver difficulty introduces substantial computational overhead because its estimation requires multiple multi-turn solver rollouts for each generated question. Together, these limitations motivate replacing solver difficulty with a training signal that can efficiently determine whether the full evidence context increases the likelihood of the target answer relative to shortcut-only contexts.

To address these limitations, we introduce \methodname{}, which changes both how training questions are constructed and how the proposer is rewarded (Figure~\ref{fig:method_compar}).Whereas Dr.~Zero relies on proposer-side search to construct questions from seed passages, \methodname{} provides relational chains from Wikidata5M together with aligned Wikipedia passages as explicit inputs for question construction. The proposer therefore starts from a predefined evidence structure and generates questions without online search. In contrast to difficulty-based rewards, which evaluate questions according to the performance of the current solver, our reward measures how much the full evidence context increases the likelihood of the target answer relative to the strongest evaluated shortcut context, including closed-book, source-only, one-shot retrieval, and individual-passage contexts. This formulation shifts proposer optimization from solver-relative difficulty to evidence dependence under explicit shortcut comparisons. Importantly, the proposed objective completely replaces rollout-based difficulty rather than supplementing it with an additional reward signal. By computing the likelihood of a fixed target answer under teacher forcing, \methodname{} eliminates solver rollouts from proposer reward estimation. Combined with structured question construction, this design substantially reduces the training time of the proposer per update.

As shown in Figure~\ref{fig:teaser}, \methodname{} improves both effectiveness and efficiency. Across five backbone models, it improves the average EM on multi-hop benchmarks over Dr.~Zero by $1.7$--$9.5$ points while accelerating proposer training by $7.37\times$--$9.61\times$ per update. Our contributions are as follows.
\begin{itemize}
    \item We identify the mismatch problem between difficulty and evidence necessity and document severe pseudo-multi-hop degeneration in a representative self-evolving search agent.
\item We introduce \methodname{}, featuring a rollout-free, shortcut-aware information-gain reward that directly measures the dependence of generated questions on the full evidence relative to evaluated shortcut contexts, promoting the construction of genuine multi-hop questions.
    \item Across seven QA benchmarks, \methodname{} improves multi-hop accuracy while reducing the proposer training time by more than $7\times$. Controlled ablations further confirm the benefits of structured question construction and the information-gain-based reward.
\end{itemize}

\section{Related Work}

\subsection{Self-evolving task generation}
Self-evolving methods reduce reliance on human-written training examples through
iterative self-training or proposer-solver self-play
\citep{zhang2025evolvesearch,li2026se}. Absolute Zero
\citep{zhao2026absolute} established zero-data self-play with an
execution-based oracle, while R-Zero \citep{huang2025r} decoupled the proposer
and solver to support broader reasoning tasks. In the search setting, SSP
\citep{lu2025search} generates questions from answers sampled from public QA
datasets and filters them for evidence sufficiency. Dr.\ Zero
\citep{yue2026dr} further removes curated QA pairs by generating questions from
corpus passages, but continues to use solver pass rate as the proposer
difficulty signal.
Recent methods introduce stronger structural and verification signals.
KG Paths \citep{wu2026knowledge} uses Wikidata subgraphs for question
construction and intermediate solver supervision, CoEvoKG
\citep{li2026coevokg} maintains an evolving KG memory, and EVE-Agent
\citep{arai2026eve} augments difficulty with evidence utility. Related analysis
also emphasizes the role of data gating in controlling which self-generated
examples enter training \citep{pu2026survive}. However, these proposer-based
methods retain solver-relative difficulty in their objectives. \methodname{} instead
constructs questions from KG chains and evaluates them using teacher-forced
full-versus-shortcut likelihood contrasts.  As shown in Figure~\ref{fig:method_compar}, this removes difficulty-based
proposer rewards and requires neither annotated QA pairs nor solver rollouts for proposer reward computation.

\subsection{Structured multi-hop construction and evidence verification}

Apparent compositionality does not guarantee genuine multi-hop dependence.
Both benchmark and synthetic questions can often be answered from a single
passage or through disconnected reasoning
\citep{jiang2019avoiding,trivedi2020multihop,
guo2023counterfactual,chen2025essential}. Prior work addresses this problem
through connected question construction
\citep{trivedi2022musique,lu2025deepdive,zhang2026searchgym},
graph-derived intermediate supervision \citep{wu2026knowledge}, or
evidence-aware verification and rewards
\citep{lu2025search,arai2026eve,zhang2026chaining}.
These mechanisms establish evidence structure, sufficiency, or utility but do not explicitly compare target-answer support under the full context against specific shortcut contexts. \methodname{} makes this comparison explicit: it scores a fixed
answer under the full chain and contrasts it with closed-book, source-only,
one-shot, and individual-passage contexts. The same knowledge-graph chain
therefore defines both the question structure and the contexts
used for shortcut-aware evaluation.

\begin{figure*}[t]
\centering
\includegraphics[width=\columnwidth]{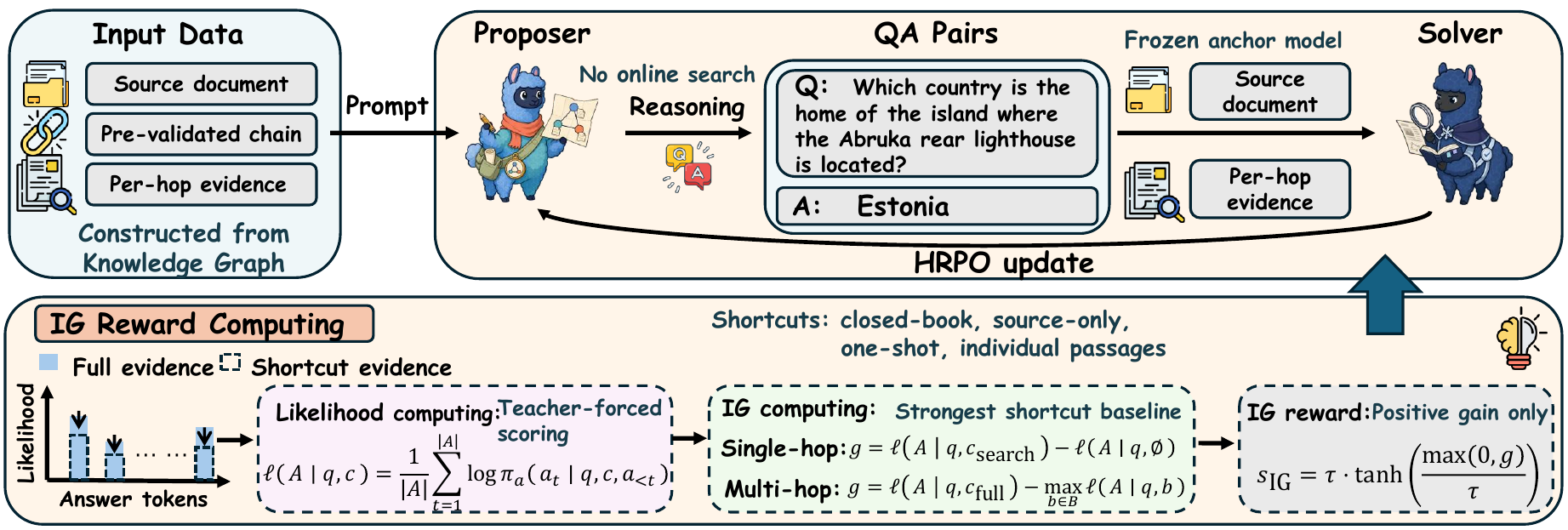}
\caption{Overview of \methodname{}. We first sample pre-validated relational chains from a knowledge graph and align each hop with supporting evidence, providing an explicit multi-hop structure for question construction. Given the validated chain and its aligned evidences, the proposer generates a question without performing online multi-turn search. The shortcut-aware information-gain reward then evaluates the generated question by comparing the likelihood of the target answer under the full evidence context against that under the strongest evaluated shortcut context, directly rewarding questions that require cross-passage evidence.}

\label{fig:overview}

\end{figure*}
\section{Method}

We adopt the proposer-solver paradigm for data-free self-evolution~\citep{yue2026dr}. A proposer $\pi_\theta$ synthesizes question-answer pairs $(x, y)$, while a solver $\pi_\phi$ learns to answer the generated questions. Both models are initialized from the same base LLM and optimized through an alternating procedure without annotated data. The two models maximize their respective expected rewards:
\begin{align}
    \text{Proposer:}\quad & \mathbb{E}_{(x,y)\sim\pi_\theta(\cdot\mid\mathcal{R})}\big[\, r(x, y) \,\big], \\
    \text{Solver:}\quad & \mathbb{E}_{(x,y)\sim\pi_\theta(\cdot\mid\mathcal{R}),\,\hat{y}\sim\pi_\phi(\cdot\mid x,\mathcal{R})}\big[\, \mathbb{I}(y = \hat{y}) \,\big],
\end{align}
where $r$ denotes the proposer reward, and $\mathbb{I}$ denotes the indicator function used for the solver reward.
In the absence of annotated data, every question-answer pair available to the solver is generated by the proposer. The central challenge is therefore to synthesize pairs that provide effective learning signals. Accordingly, we adopt the solver as given and use the standard outcome-reward formulation~\citep{jin2025searchr1trainingllmsreason,yue2026dr}. As shown in Figure~\ref{fig:overview}, we focus on the two proposer components that determine data quality: the information on which the proposer conditions (Section~\ref{sec:kg_sampler}) and the reward used to optimize the proposer (Section~\ref{sec:ig_reward}).
Together, these two components address complementary aspects of proposer optimization. The knowledge-graph-based construction provides an explicit multi-hop evidence structure for question generation, while the information-gain reward evaluates whether the full evidence provides additional support beyond the evaluated shortcut contexts.

\subsection{Knowledge-Graph Chain Sampler}
\label{sec:kg_sampler}
In Dr.\ Zero, the proposer is expected to retrieve intermediate evidence starting from a seed passage and compose the retrieved information into a multi-hop question. In practice, however, this process often collapses into single-passage question generation. Across two backbone models, most question-answer pairs labeled as multi-hop are generated without search, and the target answer is already present in the seed passage (see Sec.\ref{sec:analysis}). As a result, the assigned hop count just reflect the intended generation format rather than actual cross-document reasoning. Moreover, because pass-rate-based difficulty measures whether the current solver can answer a question rather than whether the supporting passages form a necessary and coherent evidence chain, it cannot reliably distinguish genuine multi-hop questions from such shortcut cases.

Building on the use of knowledge-graph structures to guide question construction~\citep{wu2026knowledge}, \methodname{} makes the intended evidence structure an explicit input to the proposer rather than relying on proposer-side search to discover it implicitly. We represent a relational chain as $\mathcal{C}=(e_0,r_1,e_1,\ldots,r_h,e_h)$, where each relation $r_i$ connects $e_{i-1}$ to $e_i$, $e_0$ is the source entity, and $e_h$ is the target answer. Each entity $e_i$ is further associated with an aligned Wikipedia passage. We sample such chains from Wikidata5M~\citep{wang2021kepler} through random walks over the subgraph induced by entities covered by Wiki-18.

To obtain chains with unambiguous targets and retrievable intermediate entities, we apply three filters. First, we remove cyclic paths so that each hop introduces a new entity. Second, we retain relations for which a source entity maps to a unique or nearly unique target and discard entities with ambiguous surface forms, thereby reducing the likelihood of multiple valid answers. Third, we retain a chain only when every entity $e_i$ is aligned with at least one Wiki-18 passage and at least one passage aligned with the source entity $e_0$ mentions the first bridge entity $e_1$. These filters jointly reduce entity ambiguity and enforce basic retrieval constraints, yielding structured inputs for question generation. Given a retained chain and its aligned passages, the proposer generates a question whose target answer is $e_h$, without performing multi-turn search during question construction. The information-gain reward is then applied after generation to determine whether the full evidence context supports the target answer more strongly than the evaluated shortcut contexts. Additional details of the knowledge-graph construction are provided in Appendix~\ref{app:sampler}.

\subsection{Shortcut-Aware Information-Gain Reward}
\label{sec:ig_reward}

Given a structured evidence chain, the proposer should favor question-answer pairs whose intended evidence contributes information that cannot be obtained from shortcut contexts alone. We operationalize this principle with a rollout-free, shortcut-aware information-gain reward. For a fixed question and target answer, a frozen anchor model evaluates the likelihood~\citep{wang2025infogain,liang2026ig,liang2026search} of the target answer under the full intended evidence and under each evaluated shortcut context. The reward is defined as the margin between the target-answer likelihood under the full evidence and the maximum likelihood across all shortcut contexts. A positive margin therefore indicates that the full evidence provides additional support for the target answer beyond any evaluated shortcut. We describe the reward formulation in detail below.

\textbf{Answer likelihood.}
Let $q$ denote the question generated by the proposer, and let $A=(a_1,\dots,a_{|A|})$ denote the gold answer to $q$, represented as a sequence of $|A|$ tokens. Let $c$ denote a context containing zero or more evidence passages. We define a frozen anchor model $\pi_{\text{a}}$, initialized as the base LLM in the first iteration and updated to the solver from the immediately preceding iteration thereafter. Given $q$ and $c$, the anchor model computes the length-normalized log-likelihood of the target answer:
\begin{equation}
    \ell(A \mid q, c)
    \;=\;
    \frac{1}{|A|}
    \sum_{t=1}^{|A|}
    \log \pi_{\text{a}}
    \big(a_t \mid q,\, c,\, a_{<t}\big),
\end{equation}
where $a_t$ is the $t$-th answer token and $a_{<t}$ denotes all preceding answer tokens. Normalization by the answer length $|A|$ makes the scores comparable across answers of different lengths. Computing $\ell(A \mid q,c)$ requires only a single teacher-forced forward pass over the fixed answer $A$. Unlike the pass rate, this computation does not require sampling an answer from $\pi_{\text{a}}$, thereby avoiding the repeated solver rollouts that make the baseline reward expensive to compute. Across all evidence comparisons below, the question $q$ and answer $A$ are held fixed; only the evidence context $c$ is varied.

\textbf{Single-hop questions.}
The simplest case is a single-hop chain, in which the question should be answerable from the retrieved evidence but not from parametric knowledge alone. We retrieve using the question itself,
$c_{\text{search}} = \operatorname{retrieve}(q)$, and measure how much the retrieved evidence raises the likelihood of the answer conditioned on the same question relative to the closed-book context,
\begin{equation}
    g
    \;=\;
    \ell(A \mid q, c_{\text{search}})
    -
    \ell(A \mid q, \varnothing).
\end{equation}
A large gain indicates that the retrieved passage substantially increases the likelihood assigned by the anchor model to the target answer relative to the closed-book context. In contrast, a small gain indicates either that the answer already has a high closed-book likelihood or that the retrieved passage provides little additional support.

\textbf{Multi-hop questions.}
Multi-hop questions must additionally resist predefined shortcuts that allow the answer to be recovered without cross-passage composition. Here the full-evidence context is the concatenation of all passages along the chain, denoted $c_{\text{full}}$, and we compare it against a set $B$ of shortcut contexts, each corresponding to one way in which the chain could be bypassed. For a chain with $h$ search hops, the set of shortcut contexts is

\begin{equation}
    B
    =
    \left\{
        \varnothing,\,
        c_{\text{src}},\,
        c_{\text{1-shot}},\,
        c_1,\ldots,c_h
    \right\},
    \label{eq:multihop-shortcut-set}
\end{equation}
where:
\begin{itemize}
    \item $\ell(A \mid q, \varnothing)$ is the closed-book likelihood, testing whether the question is answerable from parametric knowledge alone;
    \item $\ell(A \mid q, c_{\text{src}})$ conditions on the seed passage alone, testing whether the question can be answered without searching beyond the source;
    \item $\ell(A \mid q, c_{\text{1-shot}})$ conditions on the result of a single retrieval using $q$, testing whether the question is answerable within one retrieval hop;
    \item $\ell(A \mid q, c_k)$ conditions on the evidence passage associated with hop $k$ in isolation, testing whether any single passage suffices without cross-passage composition.
\end{itemize}
The gain is defined as the likelihood margin of the full evidence chain over the strongest shortcut, with the generated question held fixed:
\begin{equation}
    g
    \;=\;
    \ell(A \mid q, c_{\text{full}})
    -
    \max_{b \in B}
    \ell(A \mid q, b).
\end{equation}
A question therefore receives credit only when the full evidence context assigns a higher likelihood to the target answer than every evaluated shortcut context.
Finally, in both the single-hop and multi-hop cases, the gain is converted into a reward through a saturating transformation that retains only positive gains,
\begin{equation}
    s_{\text{IG}}
    \;=\;
    \tau \cdot
    \tanh\!\left(
        \frac{\max(0,\,g)}{\tau}
    \right),
\end{equation}
so that a question receives credit only when the intended evidence strictly increases the likelihood of the answer relative to the strongest applicable comparison context. The $\tanh$ function bounds the reward by $\tau= 3$ and stabilizes the reward scale across chains with different numbers of hops.

\subsection{Reward Composition and Optimization}
\label{sec:optim}

\textbf{Reward.}
We combine a graded format score with a gated information-gain reward. The format score $s_{\mathrm{fmt}}$ assigns partial credit to a parseable question-answer pair that does not reveal the answer, includes the required reasoning block, and provides an answer that matches the target entity of the chain. A separate grounding check verifies that the target answer occurs in an admissible supporting passage. For a multi-hop chain, the supporting evidence must include at least one non-seed passage from the chain rather than relying exclusively on the seed passage. The total proposer reward is

\begin{equation}
    r=\lambda s_{\mathrm{fmt}}+\mathbb{I}\!\left[s_{\mathrm{fmt}}=1\wedge
        \mathrm{grounded}
    \right]s_{\mathrm{IG}},
\end{equation}
where $\lambda$ is the format weight, set to $0.2$. Thus, the information-gain reward contributes only when a pair receives the full format score and passes the grounding check; all other pairs receive only the graded format reward. Appendix~\ref{app:reward} provides the detailed parsing, scoring, and grounding criteria.

\textbf{Optimization.} We train the proposer with hop-grouped relative policy optimization~\citep{yue2026dr} (HRPO), which standardizes the advantage within each hop group so that questions compete only against those of the same hop count,
\begin{equation}
    A_{i,h} \;=\; \frac{r_i - \operatorname{mean}_{j \in \mathcal{I}_h}(r_j)}{\operatorname{std}_{j \in \mathcal{I}_h}(r_j) + \delta},
\end{equation}
where $\mathcal{I}_h=\{j:\operatorname{hop}(j)=h\}$ is the index set of samples with hop count $h$, and $\delta$ is a small constant that ensures numerical stability.

\textbf{Solver.}
We train the solver as in prior work using GRPO~\citep{shao2024deepseekmath} with an outcome reward based on answer correctness. The proposer and solver are updated alternately. At each iteration, the proposer is updated and generates a batch of question--answer pairs, after which the solver is trained on the generated batch. The resulting solver then serves as the anchor model for the next iteration.

\section{Experiments}

\subsection{Experimental Setup}
\label{sec:setup}

\textbf{Training.}
Training proceeds for three self-evolution iterations, each comprising $50$ proposer updates, data generation, and $50$ solver updates. In each iteration, we uniformly sample one-, two-, and three-hop chains from a pool of approximately $100{,}000$ verified chains, using only the number required by the training budget rather than the entire pool. The IG anchor is the frozen base model in the first iteration and the solver from the preceding iteration thereafter. Complete training and sampling configurations are provided in Appendix~\ref{sec:implementation-details}.

\textbf{Datasets and metrics.}
We evaluate on three single-hop benchmarks---Natural Questions~\citep{kwiatkowski2019natural}, TriviaQA~\citep{joshi2017triviaqa}, and PopQA~\citep{mallen2023not}---and four multi-hop benchmarks---HotpotQA~\citep{yang2018hotpotqa}, 2WikiMultihopQA~\citep{ho2020constructing}, MuSiQue~\citep{trivedi2022musique}, and Bamboogle~\citep{press2023measuring}. We use the Search-R1 test splits~\citep{jin2025searchr1trainingllmsreason}, totaling $51{,}713$ questions, and we report exact match (EM) in Table~\ref{tab_datafree_compare}, more details are provided in Appendix~\ref{sec:More Ablations}.

\begin{table*}[!t]
    \centering
    \caption{
Performance on seven open-domain QA benchmarks measured by Exact Match (EM). For each model size, methods are grouped as few-shot/supervised or data-free. Baseline results are taken from the original papers under their reported setups.
    }
    \resizebox{\linewidth}{!}{
    \begin{tabular}{lcccccccc}
        \toprule
        Methods & NQ & TriviaQA & PopQA & HotpotQA & 2Wiki
                & MuSiQue & Bamboogle & Avg. \\
        \midrule

        \multicolumn{9}{c}{\textit{Qwen2.5-3B-Instruct}} \\
        \midrule
        \multicolumn{9}{l}{\textit{Few-shot / Supervised}} \\

        \quad IRCoT \citep{trivedi2023interleaving}
            & 0.111 & 0.312 & 0.200 & 0.164
            & 0.171 & 0.067 & 0.240 & 0.181 \\

        \quad Search-o1 \citep{li2025search}
            & 0.238 & 0.472 & 0.262 & 0.221
            & 0.218 & 0.054 & 0.320 & 0.255 \\

        \quad SFT
            & 0.249 & 0.292 & 0.104 & 0.186
            & 0.248 & 0.044 & 0.112 & 0.176 \\

        \quad R1-Instruct \citep{guo2025deepseek}
            & 0.210 & 0.449 & 0.171 & 0.208
            & 0.275 & 0.060 & 0.192 & 0.224 \\

        \quad Search-R1-Instruct
            \citep{jin2025searchr1trainingllmsreason}
            & 0.341 & 0.545 & 0.378 & 0.324
            & 0.319 & 0.103 & 0.264 & 0.325 \\

        \midrule
        \multicolumn{9}{l}{\textit{Data-Free}} \\

        \quad SQLM~\citep{chen2025self}
            & 0.264 & 0.432 & 0.258 & 0.226
            & 0.238 & 0.060 & 0.158 & 0.233 \\

        \quad R-Zero~\citep{huang2025r}
            & 0.389 & 0.513 & 0.370 & 0.243
            & 0.128 & 0.052 & 0.096 & 0.256 \\

        \quad Dr.\ Zero \citep{yue2026dr}
            & \textbf{0.397} & \textbf{0.572}
            & \textbf{0.431} & 0.298
            & 0.291 & 0.091 & 0.200 & 0.326 \\

        \rowcolor{gray!25}
        \quad \textbf{\methodname} (Ours)
            & 0.380 & \textbf{0.572}
            & 0.416 & \textbf{0.353}
            & \textbf{0.390} & \textbf{0.172}
            & \textbf{0.344} & \textbf{0.375} \\

        \midrule
        \multicolumn{9}{c}{\textit{Qwen2.5-7B-Instruct}} \\
        \midrule
        \multicolumn{9}{l}{\textit{Few-shot / Supervised}} \\

        \quad IRCoT \citep{trivedi2023interleaving}
            & 0.224 & 0.478 & 0.301 & 0.133
            & 0.149 & 0.072 & 0.224 & 0.239 \\

        \quad Search-o1 \citep{li2025search}
            & 0.151 & 0.443 & 0.131 & 0.187
            & 0.176 & 0.058 & 0.296 & 0.206 \\

        \quad SFT
            & 0.318 & 0.354 & 0.121 & 0.217
            & 0.259 & 0.066 & 0.112 & 0.207 \\

        \quad R1-Instruct \citep{guo2025deepseek}
            & 0.270 & 0.537 & 0.199 & 0.237
            & 0.292 & 0.072 & 0.293 & 0.271 \\

        \quad Search-R1-Instruct
            \citep{jin2025searchr1trainingllmsreason}
            & 0.393 & \textbf{0.610}
            & 0.397 & 0.370
            & \textbf{0.414} & 0.146
            & 0.368 & 0.385 \\

        \midrule
        \multicolumn{9}{l}{\textit{Data-Free}} \\

        \quad SSP \citep{lu2025search}
            & 0.354 & 0.576 & 0.428 & 0.322
            & 0.304 & 0.154 & 0.344 & 0.355 \\

        \quad KG Paths \citep{wu2026knowledge}
            & 0.352 & 0.568 & 0.408 & 0.330
            & 0.322 & 0.128 & 0.400 & 0.358 \\

        \quad Dr.\ Zero \citep{yue2026dr}
            & \textbf{0.406} & 0.608
            & 0.416 & 0.362
            & 0.347 & 0.104 & 0.360 & 0.372 \\

        \rowcolor{gray!25}
        \quad \textbf{\methodname} (Ours)
            & 0.397 & \textbf{0.610}
            & \textbf{0.438} & \textbf{0.387}
            & 0.389 & \textbf{0.164}
            & \textbf{0.424} & \textbf{0.401} \\

        \bottomrule
    \end{tabular}
    }

    \label{tab_datafree_compare}
\end{table*}

\subsection{Main Results}
\label{sec:main_results}

Table~\ref{tab_datafree_compare} reports exact-match performance on seven benchmarks. \methodname{} achieves the highest average performance at both model scales, outperforming Dr.\ Zero and the matched Search-R1-Instruct baseline. The improvements are most pronounced on multi-hop QA, where \methodname{} leads on nearly all benchmarks across both scales. This pattern aligns with the shortcut analysis in Table~\ref{tab:diag} and suggests that our explicit evidence chains and the IG reward are particularly beneficial for multi-hop reasoning. More results of \methodname{} on different model scales are shown in Table~\ref{tab:dir-free-main-results} in the Appendix.

\subsection{Ablation Study}
\label{sec:ablation}

\begin{table*}[!t]
    \centering
    \caption{
        Ablation of \methodname \ with Qwen2.5-3B-Instruct.
        \emph{Difficulty Only} replaces the IG reward with the
        pass-rate difficulty reward;
        \emph{IG + Difficulty} combines both rewards;
        \emph{w/o KG} restores Dr.\ Zero-style proposer search; and
        \emph{w/o Proposer Training} uses the base proposer to generate
        questions from KG chains without reward-based optimization.
        Best results are shown in bold.
    }
    \vspace{0.2cm}
    \resizebox{\linewidth}{!}{
    \begin{tabular}{lcccccccc}
        \toprule
        & NQ & TriviaQA & PopQA & HotpotQA
        & 2Wiki & MuSiQue & Bamboogle & Avg. \\
        \midrule

        \textbf{\methodname}
        & \textbf{0.380}
        & \textbf{0.572}
        & \textbf{0.416}
        & \textbf{0.353}
        & \textbf{0.390}
        & \textbf{0.172}
        & \textbf{0.344}
        & \textbf{0.375} \\

        \quad Difficulty Only
        & 0.362
        & 0.568
        & 0.405
        & 0.324
        & 0.357
        & 0.121
        & 0.264
        & 0.343 \\

        \quad IG + Difficulty
        & 0.358
        & 0.563
        & 0.389
        & 0.325
        & 0.357
        & 0.137
        & 0.248
        & 0.340 \\

        \quad w/o KG
        & 0.366
        & 0.536
        & 0.365
        & 0.319
        & 0.356
        & 0.129
        & 0.272
        & 0.335 \\

        \quad w/o Proposer Training
        & 0.377
        & 0.570
        & 0.393
        & 0.289
        & 0.281
        & 0.063
        & 0.168
        & 0.306 \\

        \bottomrule
    \end{tabular}
    }

    \label{tab:ablation}
\end{table*}

\textbf{Module Ablation.}
Table~\ref{tab:ablation} evaluates the contribution of the KG chain sampler and the proposer reward. Removing proposer optimization causes the largest performance drop, reducing the average score from $0.375$ to $0.306$, with most of the degradation occurring on multi-hop QA benchmarks. This result indicates that structured evidence chains alone are insufficient to produce effective training questions without reward-guided proposer optimization. Replacing the information-gain (IG) reward with the difficulty reward reduces the average score to $0.343$, while combining the two yields a similar score of $0.340$. These results suggest that, under our setting, the evaluated difficulty signal neither provides an effective substitute for the IG reward nor offers additional benefit when combined with it. Removing the KG scaffold also reduces the average score to $0.335$, demonstrating the importance of explicit relational structure for question construction. Overall, both components contribute to the performance of \methodname{}, with IG-based proposer optimization accounting for the largest improvement.

\textbf{Shortcut Ablation.}
\begin{table*}[!t]
    \centering
\caption{Leave-one-shortcut-type-out ablation of the shortcut set ${B}$. Each row removes one type of shortcut context from the full set. All four shortcut types contribute positively on average, with the largest degradation occurring when the source-only hop passage contexts are removed.}
    \resizebox{\linewidth}{!}{
    \begin{tabular}{lcccccccc}
        \toprule
        & NQ & TriviaQA & PopQA & HotpotQA & 2Wiki & MuSiQue & Bamboogle & Avg. \\
        \midrule
        \textbf{\methodname} (full $B$)
        & \textbf{0.380}
        & \textbf{0.572}
        & \textbf{0.416}
        & \textbf{0.353}
        & \textbf{0.390}
        & \textbf{0.172}
        & 0.344
        & \textbf{0.375} \\
        \quad w/o one-shot
        & 0.366
        & 0.564
        & 0.390
        & 0.324
        & 0.373
        & 0.142
        & \textbf{0.360}
        & 0.360 \\
        \quad w/o closed-book
        & 0.341
        & 0.556
        & 0.370
        & 0.315
        & 0.326
        & 0.117
        & 0.288
        & 0.330 \\
        \quad w/o per-hop
        & 0.321
        & 0.535
        & 0.355
        & 0.290
        & 0.311
        & 0.114
        & 0.320
        & 0.321 \\
        \quad w/o source-only
        & 0.326
        & 0.527
        & 0.356
        & 0.295
        & 0.303
        & 0.104
        & 0.304
        & 0.316 \\
        \bottomrule
    \end{tabular}
    }
    \label{tab:shortcut}
\end{table*}

Table~\ref{tab:shortcut} removes each shortcut context from $B$ individually. Every removal reduces average downstream performance, indicating that the shortcut contexts provide complementary training signals. The source-only shortcut has the largest effect, reducing the average from $0.375$ to $0.316$ ($-0.059$). Without this comparison, the reward no longer directly penalizes cases in which the seed passage alone assigns high likelihood to the target answer. Removing the per-hop shortcut yields a similarly large drop to $0.321$ ($-0.054$), because the reward no longer contrasts the full-context likelihood against each individual evidence passage. Removing the closed-book and one-shot shortcuts lowers the average to $0.330$ ($-0.045$) and $0.360$ ($-0.015$), respectively. The reductions are generally larger on the multi-hop benchmarks, suggesting that the complete shortcut set helps construct training questions that improve downstream multi-hop QA performance. Concrete reward traces illustrating how the source-only and per-hop contexts prevent false-positive rewards are provided in Appendix~\ref{app:shortcut-cases}.

\subsection{Analysis of Proposer Search and Shortcut Behavior}
\label{sec:analysis}

\begin{table*}[t]
  \centering
  \small
  \caption{Behavior of the Dr.~Zero proposer under two backbone
  replications. Across both models, most questions labeled as multi-hop are
  generated without any search, and their answers are frequently recoverable
  from the seed passage.}
  \begin{minipage}[t]{0.485\textwidth}
    \centering
    \textbf{(a) Qwen2.5-3B-Instruct}
    \vspace{3pt}

    \small
    \setlength{\tabcolsep}{2.5pt}
    \renewcommand{\arraystretch}{1.05}
    \begin{tabular}{@{}lrrr@{}}
      \toprule
      Hop label & \# Pairs & Zero-search & Answer in seed \\
      \midrule
      1 & 13{,}722 & --- & 96.9\% \\
      2 & 10{,}131 & 100.0\% & 96.4\% \\
      3 & 6{,}630 & 99.98\% & 96.3\% \\
      4 & 3{,}440 & 100.0\% & 96.0\% \\
      \midrule
      $\geq 2$ & \textbf{20{,}201} & \textbf{99.995\%} & \textbf{96.4\%} \\
      \bottomrule
    \end{tabular}
  \end{minipage}
  \hfill
  \begin{minipage}[t]{0.485\textwidth}
    \centering
    \textbf{(b) Qwen3-4B}
    \vspace{3pt}

    \small
    \setlength{\tabcolsep}{2.5pt}
    \renewcommand{\arraystretch}{1.05}
    \begin{tabular}{@{}lrrr@{}}
      \toprule
      Hop label & \# Pairs & Zero-search & Answer in seed \\
      \midrule
      1 & 13{,}602 & --- & 93.2\% \\
      2 & 10{,}192 & 79.34\% & 83.1\% \\
      3 & 6{,}791 & 90.21\% & 85.4\% \\
      4 & 3{,}338 & 90.05\% & 83.1\% \\
      \midrule
      $\geq 2$ & \textbf{20{,}321} & \textbf{84.73\%} & \textbf{83.9\%} \\
      \bottomrule
    \end{tabular}
  \end{minipage}
  \label{tab:diag}
\end{table*}

Prior data-free proposers may generate questions with pseudo-multi-hop labels without performing genuine multi-document search. To determine whether this behavior persists across backbones, we reproduce Dr.~Zero~\citep{yue2026dr} with Qwen2.5-3B-Instruct and Qwen3-4B under the same retrieval configuration using E5-base-v2 and Wiki-18. Each replication generates $33{,}923$ question--answer pairs for solver training.

Table~\ref{tab:diag} shows that the shortcut occurs with both backbones. Among questions labeled as multi-hop, the zero-search rates are $99.995\%$ for Qwen2.5-3B-Instruct and $84.73\%$ for Qwen3-4B, while $96.4\%$ and $83.9\%$ of the corresponding answers occur in the seed passages. The stronger backbone reduces but does not eliminate the shortcut. Representative cases in Appendix~\ref{app:drzero-cases} further illustrate that the hop label often reflects the requested output format rather than actual cross-document composition. These findings provide a plausible explanation for the limited multi-hop gains of Dr.~Zero.

\begin{table*}[t]
  \centering
  \caption{Gold-blind counterfactual audit of four observable reasoning
  shortcuts. Each method contributes 100 questions with a matched 51/49
  two-/three-hop distribution. Brackets report Wilson 95\% confidence
  intervals. The final two rows report one-sided Fisher exact-test
  $p$-values for \methodname.}
  \label{tab:shortcut_audit}
  \footnotesize
  \setlength{\tabcolsep}{4pt}
  \renewcommand{\arraystretch}{1.08}
  \begin{tabular*}{\textwidth}{@{\extracolsep{\fill}}lcccc@{}}
    \toprule
    \textbf{Method}
      & \makecell[c]{\textbf{Closed-book}\\\textbf{shortcut} $\downarrow$}
      & \makecell[c]{\textbf{Source-only}\\\textbf{shortcut} $\downarrow$}
      & \makecell[c]{\textbf{One-shot}\\\textbf{shortcut} $\downarrow$}
      & \makecell[c]{\textbf{Single-component}\\\textbf{shortcut} $\downarrow$} \\
    \midrule
    Dr.\ Zero
      & \makecell[c]{17\%\\[-1pt]
          {\scriptsize\color{black!60}[10.9, 25.5]}}
      & \makecell[c]{81\%\\[-1pt]
          {\scriptsize\color{black!60}[72.2, 87.5]}}
      & \makecell[c]{65\%\\[-1pt]
          {\scriptsize\color{black!60}[55.3, 73.6]}}
      & \makecell[c]{88\%\\[-1pt]
          {\scriptsize\color{black!60}[80.2, 93.0]}} \\
    KG + Difficulty
      & \makecell[c]{48\%\\[-1pt]
          {\scriptsize\color{black!60}[38.5, 57.7]}}
      & \makecell[c]{20\%\\[-1pt]
          {\scriptsize\color{black!60}[13.3, 28.9]}}
      & \makecell[c]{45\%\\[-1pt]
          {\scriptsize\color{black!60}[36.0, 55.2]}}
      & \makecell[c]{51\%\\[-1pt]
          {\scriptsize\color{black!60}[41.3, 60.6]}} \\
    \methodname
      & \makecell[c]{\textbf{5\%}\\[-1pt]
          {\scriptsize\color{black!60}[2.2, 11.2]}}
      & \makecell[c]{\textbf{3\%}\\[-1pt]
          {\scriptsize\color{black!60}[1.0, 8.5]}}
      & \makecell[c]{\textbf{2\%}\\[-1pt]
          {\scriptsize\color{black!60}[0.6, 7.0]}}
      & \makecell[c]{\textbf{5\%}\\[-1pt]
          {\scriptsize\color{black!60}[2.2, 11.2]}} \\
    \midrule
    \multicolumn{5}{@{}l}{\textit{\methodname \ vs.\ baseline ($p$-value)}} \\
    vs.\ Dr.\ Zero
      & {\scriptsize $5.7\times10^{-3}$}
      & {\scriptsize $3.1\times10^{-33}$}
      & {\scriptsize $3.8\times10^{-24}$}
      & {\scriptsize $1.4\times10^{-36}$} \\
    vs.\ KG + Difficulty
      & {\scriptsize $7.0\times10^{-13}$}
      & {\scriptsize $1.1\times10^{-4}$}
      & {\scriptsize $1.5\times10^{-14}$}
      & {\scriptsize $3.9\times10^{-14}$} \\
    \bottomrule
  \end{tabular*}
\end{table*}

We further evaluate whether \methodname{} suppresses shortcut behavior through a controlled, gold-blind counterfactual audit over the four contexts in $\mathcal{B}$: closed-book, source-only, one-shot retrieval, and individual chain components. Each method contributes 100 questions with the same two-/three-hop distribution, and each counterfactual context is evaluated through an independent judge call.

Table~\ref{tab:shortcut_audit} reveals distinct shortcut profiles across methods. For Dr.\ Zero, the source-only and single-component shortcut rates reach $81\%$ and $88\%$, respectively, whereas the closed-book shortcut rate is only $17\%$. This pattern suggests that the dominant shortcuts arise from local evidence contained in the provided passages rather than from parametric knowledge alone. In contrast, KG + Difficulty, corresponding to the \emph{Difficulty Only} ablation in Table~\ref{tab:ablation}, exhibits a substantially higher closed-book shortcut rate of $48\%$, indicating that the difficulty reward does not explicitly penalize questions that can be answered from parametric knowledge. \methodname{} reduces all four shortcut rates to between $2\%$ and $5\%$, with statistically significant improvements over both baselines in every comparison. These results align with the IG objective, which explicitly rewards a likelihood advantage of the full evidence over the strongest evaluated shortcut context.

\section{Conclusions}
We introduced \methodname{}, a self-evolving search framework that replaces solver-relative difficulty with a shortcut-aware objective for measuring evidence dependence. Across five backbone models and seven open-domain QA benchmarks, \methodname{} consistently improves overall performance, with particularly strong gains on multi-hop tasks, while reducing proposer training time by more than \(7\times\). Ablation studies verify the contributions of both structured chain construction and the shortcut-aware information-gain reward. Overall, \methodname{} provides a principled and practical approach for generating higher-quality training questions while improving both the effectiveness and efficiency of data-free search-agent self-evolution.

\bibliography{references}
\bibliographystyle{references}

\newpage
\appendix
\section{Appendix}

\subsection{Implementations}
\label{sec:implementation-details}

\begin{table*}[t]
  \centering
    \caption{Representative proposer and solver configurations for \methodname.}
    \vspace{0.2cm}
  \begin{minipage}[t]{0.485\textwidth}
    \centering
    \textbf{(a) Proposer (challenger)}
    \vspace{3pt}

    \small
    \setlength{\tabcolsep}{4pt}
    \renewcommand{\arraystretch}{1.05}
    \begin{tabularx}{\linewidth}{@{}l>{\raggedleft\arraybackslash}X@{}}
      \toprule
      Hyperparameter & Setting \\
      \midrule
      Algorithm & HRPO \\
      Maximum training steps & 50 \\
      Optimizer & AdamW \\
      Optimizer momentum & $\beta_1,\beta_2=0.9,0.999$ \\
      Learning rate & $1\times10^{-6}$ \\
      Warmup ratio & 0.03 \\
      LR scheduler & Constant with warmup \\
      Weight decay & 0.01 \\
      Maximum gradient norm & 0.1 \\
      Group size in HRPO & 1 \\
      Total train batch size & 256 \\
      PPO mini-batch size & 128 \\
      Precision & BF16-mixed \\
      Maximum prompt length & 2560 \\
      Multi-turn tool use & Disabled \\
      \bottomrule
    \end{tabularx}
  \end{minipage}
  \hfill
  \begin{minipage}[t]{0.485\textwidth}
    \centering
    \textbf{(b) Solver}
    \vspace{3pt}

    \small
    \setlength{\tabcolsep}{4pt}
    \renewcommand{\arraystretch}{1.05}
    \begin{tabularx}{\linewidth}{@{}l>{\raggedleft\arraybackslash}X@{}}
      \toprule
      Hyperparameter & Setting \\
      \midrule
      Algorithm & GRPO \\
      Maximum training steps & 50 \\
      Optimizer & AdamW \\
      Optimizer momentum & $\beta_1,\beta_2=0.9,0.999$ \\
      Learning rate & $1\times10^{-6}$ \\
      Warmup ratio & 0.03 \\
      LR scheduler & Constant with warmup \\
      Weight decay & 0.01 \\
      Maximum gradient norm & 0.1 \\
      Group size in GRPO & 5 \\
      Total train batch size & 256 \\
      PPO mini-batch size & 128 \\
      Precision & BF16-mixed \\
      Maximum prompt length & 512 \\
      Multi-turn tool use & Enabled (search tool) \\
      \bottomrule
    \end{tabularx}
  \end{minipage}

  \label{tab:dir-free-hparams}
\end{table*}

We evaluate \methodname{} across multiple backbone sizes and initialize both the proposer and solver from the same backbone. The construction pipeline produces approximately $100{,}000$ verified chains, which form a candidate pool rather than a training set used exhaustively. In each self-evolution iteration, we uniformly sample one-, two-, and three-hop chains according to the training budget. With a total batch size of $256$, the $50$ proposer updates require $12{,}800$ chain instances. Subsequent data generation is restricted to the question--answer pairs required for $50$ solver updates.

\methodname{} inherits  Dr.~Zero's Wiki-18 index, E5-base-v2 retriever, search-turn limit and retrieval top-k.
Following Dr.~Zero~\citep{yue2026dr}, we also optimize the proposer using HRPO with one generation per prompt and the information-gain reward. Proposer optimization runs for at most $50$ steps with a maximum prompt length of $2{,}560$ tokens and no tool interaction. We then train the solver on the generated question--answer pairs using GRPO~\citep{shao2024deepseekmath}. The solver samples five trajectories per question, interacts with the search tool over multiple turns, and is optimized for at most $50$ steps with a maximum prompt length of $512$ tokens.

Both stages use AdamW~\citep{loshchilov2017decoupled} with a learning rate of $1{\times}10^{-6}$, a $3\%$ warmup ratio, a weight decay of $0.01$, and a maximum gradient norm of $0.1$. We use a total training batch size of $256$, a PPO mini-batch size of $128$, a constant learning-rate schedule with warmup, and BF16 mixed precision. Table~\ref{tab:dir-free-hparams} summarizes the complete configurations.

\subsection{Proposer efficiency comparison}

\textbf{Qwen2.5-3B-Instruct.}
Under the same 8$\times$H800 setup, \methodname{} reduces the runtime of 50 proposer updates from 7.88 to 0.82 hours ($9.6\times$). Generating the complete solver-training pool decreases from 6.30 to 1.04 hours ($6.1\times$), or from 0.669 to 0.094 seconds per question ($7.1\times$). This reduction comes from replacing Dr.~Zero's live retrieval with five samples per prompt by direct question generation from a grounded KG path. Overall, proposer training and data generation decrease from 14.18 to 1.86 hours ($7.6\times$). Including 50 solver updates, data conversion, and evaluation, the estimated end-to-end runtime decreases from 22.02 to 8.61 hours ($2.56\times$), reducing GPU-hours by 60.9\%. The KG-chain pool is constructed and validated only once, taking less than one hour, and is reused in all subsequent iterations. We therefore exclude this one-time preprocessing cost from the per-iteration comparison.

\textbf{Standardized cross-backbone comparison.}
We additionally standardize data generation to the 12.8k questions required by 50 solver updates. For Qwen2.5-3B-Instruct, generation time decreases from 4.75 to 0.67 hours ($7.14\times$), while the combined proposer-side runtime decreases from 12.63 to 1.49 hours ($8.48\times$), reducing GPU-hours by 88.2\%. For Qwen2.5-7B-Instruct, proposer training decreases from 8.42 to 0.94 hours ($8.96\times$), generation from 3.93 to 0.89 hours ($4.42\times$), and the combined runtime from 12.35 to 1.83 hours ($6.75\times$), reducing GPU-hours by 85.2\%.

\textbf{Overall trend.}
Qwen3-4B/8B and Llama-3.1-8B-Instruct show the same pattern: proposer training is accelerated by $7.37\times$--$8.04\times$, data generation by $3.42\times$--$5.30\times$, and the combined proposer-side pipeline by $5.61\times$--$6.92\times$. Their per-iteration GPU-hour reductions range from 82.2\% to 85.6\%. Across all five backbones, the combined proposer-side speedup averages $6.8\times$, with no systematic dependence on model scale.

\subsection{More Ablations}
\label{sec:More Ablations}
\begin{table*}[ht]
    \centering
    \small
        \caption{Effect of the hop-sampling ratio used to construct proposer training data. All columns report EM under the same training and evaluation protocol. Values are rounded to three decimal places, and bolding is determined using the unrounded values.}
        \vspace{0.2cm}
    \resizebox{\linewidth}{!}{
    \begin{tabular}{lcccccccc}
        \toprule
        Setting
        & NQ
        & TriviaQA
        & PopQA
        & HotpotQA
        & 2Wiki
        & MuSiQue
        & Bamboogle
        & Avg. \\
        \midrule
        Ratio 1:1:1:1
        & 0.377
        & 0.570
        & 0.403
        & 0.338
        & 0.352
        & 0.141
        & 0.264
        & 0.349 \\
        Ratio 4:3:2
        & 0.375
        & 0.572
        & 0.408
        & 0.324
        & 0.354
        & 0.129
        & 0.296
        & 0.351 \\
\textbf{Ratio 1:1:1}
        & \textbf{0.380}
        & \textbf{0.572}
        & \textbf{0.416}
        & \textbf{0.353}
        & \textbf{0.390}
        & \textbf{0.172}
        & \textbf{0.344}
        & \textbf{0.375}  \\
        \bottomrule
    \end{tabular}
    }

    \label{tab:hop_ratio}
\end{table*}

\definecolor{dirfreerow}{RGB}{226,239,248}

\begin{table*}[t]
  \centering
  \caption{Evaluation with Exact Match (EM) and token-level F1. Each cell
  reports EM / F1 in percent. Boldface marks the best comparable value within
  each backbone group. F1 is compared only when both methods report it; blank
  entries denote results that have not yet been provided.}
  \vspace{0.2cm}
  \label{tab:dir-free-main-results}
  \setlength{\tabcolsep}{4.2pt}
  \renewcommand{\arraystretch}{1.00}
  \setlength{\aboverulesep}{0.5pt}
  \setlength{\belowrulesep}{1.0pt}
  \resizebox{\textwidth}{!}{%
  \begin{tabular}{lcccccccc}
    \toprule
    & \multicolumn{3}{c}{GeneralQA}
    & \multicolumn{4}{c}{Multi-HopQA} & \\
    \cmidrule(lr){2-4}\cmidrule(lr){5-8}
    Method & NQ & TriviaQA & PopQA & HotpotQA & 2Wiki & MuSiQue
      & Bamboogle & Avg. \\
    \midrule

    \multicolumn{9}{c}{\textit{Qwen2.5-3B-Instruct}} \\
    \midrule
    Dr.~Zero
      & \textbf{39.7}\,/\,\phantom{00.0}
      & \textbf{57.2}\,/\,\phantom{00.0}
      & \textbf{43.1}\,/\,\phantom{00.0}
      & 29.8\,/\,\phantom{00.0}
      & 29.1\,/\,\phantom{00.0}
      & 9.1\,/\,\phantom{00.0}
      & 20.0\,/\,\phantom{00.0}
      & 32.6\,/\,\phantom{00.0} \\
    \rowcolor{dirfreerow}
    \textbf{\methodname (Ours)}
      & 38.0\,/\,47.5
      & \textbf{57.2}\,/\,65.0
      & 41.6\,/\,45.8
      & \textbf{35.3}\,/\,46.1
      & \textbf{39.0}\,/\,45.2
      & \textbf{17.2}\,/\,25.0
      & \textbf{34.4}\,/\,47.9
      & \textbf{37.5}\,/\,46.1 \\
    \midrule

    \multicolumn{9}{c}{\textit{Qwen2.5-7B-Instruct}} \\
    \midrule
    Dr.~Zero
      & \textbf{40.6}\,/\,\phantom{00.0}
      & 60.8\,/\,\phantom{00.0}
      & 41.6\,/\,\phantom{00.0}
      & 36.2\,/\,\phantom{00.0}
      & 34.7\,/\,\phantom{00.0}
      & 10.4\,/\,\phantom{00.0}
      & 36.0\,/\,\phantom{00.0}
      & 37.2\,/\,\phantom{00.0} \\
    \rowcolor{dirfreerow}
    \textbf{\methodname (Ours)}
      & 39.7\,/\,50.1
      & \textbf{61.0}\,/\,69.0
      & \textbf{43.8}\,/\,48.5
      & \textbf{38.7}\,/\,49.8
      & \textbf{38.9}\,/\,46.2
      & \textbf{16.4}\,/\,23.5
      & \textbf{42.4}\,/\,53.0
      & \textbf{40.1}\,/\,48.6 \\
    \midrule

    \multicolumn{9}{c}{\textit{LLaMA-3.1-8B-Instruct}} \\
    \midrule
    Dr.~Zero
      & \textbf{42.9}\,/\,\textbf{52.3}
      & \textbf{64.3}\,/\,\textbf{72.3}
      & 45.2\,/\,\textbf{51.7}
      & 31.7\,/\,42.0
      & \textbf{31.4}\,/\,\textbf{37.4}
      & 7.5\,/\,14.7
      & 16.8\,/\,27.8
      & 34.2\,/\,42.6 \\
    \rowcolor{dirfreerow}
    \textbf{\methodname (Ours)}
      & 41.5\,/\,50.8
      & 62.5\,/\,70.8
      & \textbf{46.2}\,/\,49.3
      & \textbf{32.6}\,/\,\textbf{44.2}
      & 31.0\,/\,37.1
      & \textbf{12.6}\,/\,\textbf{20.7}
      & \textbf{35.2}\,/\,\textbf{44.8}
      & \textbf{37.4}\,/\,\textbf{45.4} \\
    \midrule

    \multicolumn{9}{c}{\textit{Qwen3-4B}} \\
    \midrule
    Dr.~Zero
      & 38.7\,/\,49.8
      & 61.6\,/\,69.8
      & 40.3\,/\,47.5
      & 36.2\,/\,47.2
      & 38.8\,/\,45.5
      & 14.0\,/\,22.0
      & 40.8\,/\,52.4
      & 38.6\,/\,47.7 \\
    \rowcolor{dirfreerow}
    \textbf{\methodname (Ours)}
      & \textbf{42.0}\,/\,\textbf{51.3}
      & \textbf{64.5}\,/\,\textbf{71.4}
      & \textbf{44.9}\,/\,\textbf{49.8}
      & \textbf{38.4}\,/\,\textbf{49.0}
      & \textbf{41.1}\,/\,\textbf{46.5}
      & \textbf{15.5}\,/\,\textbf{23.4}
      & \textbf{41.6}\,/\,\textbf{53.0}
      & \textbf{41.1}\,/\,\textbf{49.2} \\
    \midrule

    \multicolumn{9}{c}{\textit{Qwen3-8B}} \\
    \midrule
    Dr.~Zero
      & 39.5\,/\,51.3
      & 65.2\,/\,73.1
      & 42.0\,/\,48.7
      & 38.5\,/\,49.9
      & 42.8\,/\,49.1
      & 13.5\,/\,21.9
      & 41.6\,/\,53.4
      & 40.4\,/\,49.6 \\
    \rowcolor{dirfreerow}
    \textbf{\methodname (Ours)}
      & \textbf{39.9}\,/\,\textbf{51.4}
      & \textbf{66.7}\,/\,\textbf{74.1}
      & \textbf{43.8}\,/\,\textbf{49.7}
      & \textbf{40.3}\,/\,\textbf{51.8}
      & \textbf{46.5}\,/\,\textbf{52.1}
      & \textbf{16.6}\,/\,\textbf{24.8}
      & \textbf{45.6}\,/\,\textbf{60.5}
      & \textbf{42.8}\,/\,\textbf{52.0} \\

    \bottomrule
  \end{tabular}%
  }
\end{table*}

\subsubsection{Ablations on hop ratio}
Table~\ref{tab:hop_ratio} examines how the hop composition of the proposer training data affects downstream performance. Ratio $1{:}1{:}1$, which assigns equal proportions to one-, two-, and three-hop examples, achieves the highest EM on all seven benchmarks. Compared with Ratio $4{:}3{:}2$, increasing the relative proportion of multi-hop examples produces particularly clear gains on HotpotQA, 2Wiki, MuSiQue, and Bamboogle, demonstrating the effectiveness of \methodname{} in constructing useful multi-hop supervision. However, extending the mixture to four-hop examples with Ratio $1{:}1{:}1{:}1$ reduces the average EM from 0.375 to 0.349. Constructing well-posed questions with unambiguous intermediate references and complete evidence dependencies becomes increasingly difficult as chain length grows. The performance degradation may therefore reflect the lower reliability of the four-hop question--answer pairs. We accordingly adopt Ratio $1{:}1{:}1$ as the default configuration.

\subsubsection{ Performance across model scales}
Table~\ref{tab:dir-free-main-results} compares \methodname{} with Dr.~Zero across five model backbones and seven QA benchmarks. \methodname{} consistently achieves higher average EM across all backbones and improves average F1 whenever comparable results are available. The gains are particularly consistent on multi-hop benchmarks such as HotpotQA, MuSiQue, and Bamboogle, while Dr.~Zero remains competitive on several single-hop datasets. Overall, these results demonstrate that the effectiveness of \methodname{} generalizes across model families and scales, with its main advantage arising from improved performance on multi-hop QA benchmarks.

\subsubsection{Effect of the Anchor-Model Update Strategy}
\label{app:anchor-update}

\begin{figure}[t]
    \centering
    \includegraphics[width=\columnwidth]{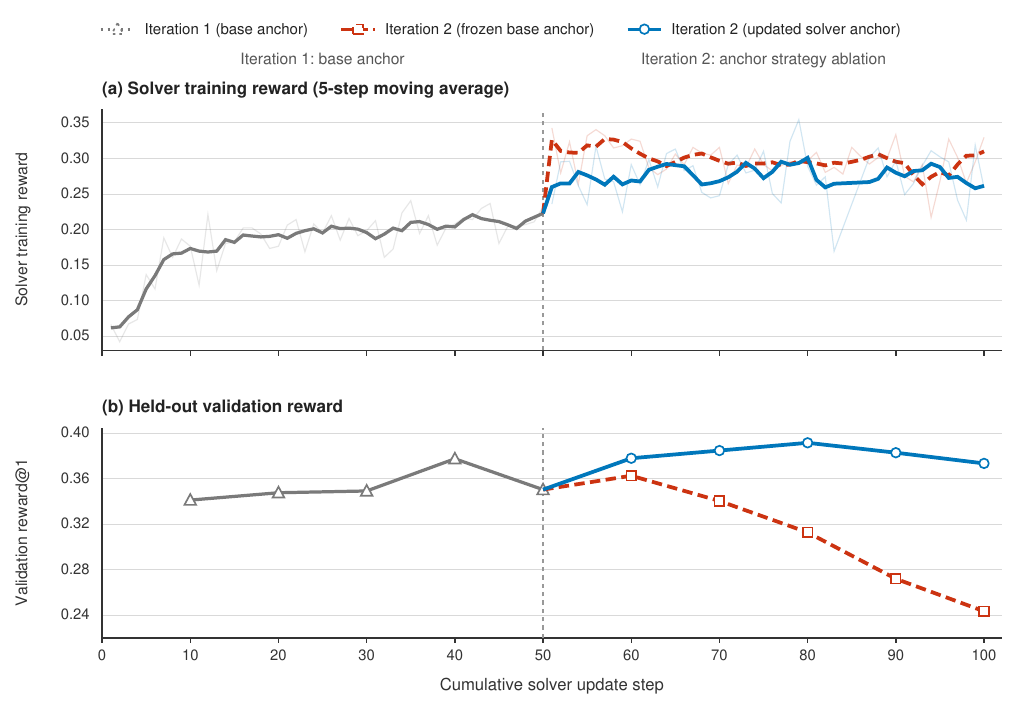}
    \caption{Anchor-update ablation across two iterations. Panel~(a) shows the solver's raw training rewards (faint) and five-step moving averages; Panel~(b) shows held-out validation reward. Iteration~1 is the shared prefix, and the two Iteration-2 strategies branch at cumulative step 50. The isolated excursion at updated-anchor steps 34--36 is omitted from the plot but retained in the accompanying data.}
    \label{fig:anchor-update-serial}
\end{figure}

We ablate the default anchor update by retaining the base-model anchor in Iteration~2 while keeping all other settings fixed. Figure~\ref{fig:anchor-update-serial} plots Iteration~1 once over steps 1--50 and places Iteration~2 step $k$ at cumulative step $50+k$. The frozen-anchor solver achieves a higher training reward, but its validation reward drops sharply from 0.3626 to 0.2434. In contrast, the updated-anchor solver attains a lower training reward while largely preserving validation performance, from 0.3780 to 0.3734, and reaches a peak of 0.3915. Because Panel~(a) reports the solver reward rather than the proposer IG reward, this discrepancy cannot be attributed to anchor-dependent reward calibration. Instead, the train--validation divergence is consistent with a distributional effect induced by proposer optimization: the proposer has already been trained to favor questions that receive high reward under the fixed anchor, making the resulting training questions easier for that same anchor and therefore yielding inflated training rewards without comparable gains in held-out performance. Updating the anchor changes this reference model and consequently shifts proposer optimization toward questions that remain informative for a stronger solver. In Iteration~3, the anchor is further updated to the Iteration~2 solver. The continued improvement in held-out performance across all three iterations (Table~\ref{tab:iters}) indicates that, at this training scale, progressively updating the anchor does not cause IG signal compression to accumulate into reward collapse.

\subsection{Effect of Self-Evolution Iterations}
\label{sec:iterations}
\begin{table*}[!ht]
    \centering
    \caption{Performance of \methodname \ across self-evolution iterations. The best result in each column within each backbone is shown in bold. Both backbones improve across iterations, with the largest gains occurring on multi-hop benchmarks.}
    \vspace{0.2cm}
    \resizebox{\linewidth}{!}{
    \begin{tabular}{lcccccccc}
        \toprule
        & NQ & TriviaQA & PopQA & HotpotQA & 2Wiki & MuSiQue & Bamboogle & Avg. \\
        \midrule
        \multicolumn{9}{c}{\textit{Qwen2.5-3B-Instruct}} \\
        \midrule
        \methodname \ Iter 1
        & 0.358 & 0.563 & 0.399 & 0.336
        & 0.364 & 0.138 & 0.320 & 0.354 \\
        \methodname \ Iter 2
        & 0.365 & 0.565 & 0.413 & 0.344
        & 0.379 & 0.141 & 0.320 & 0.362 \\
        \methodname \ Iter 3
        & \textbf{0.380} & \textbf{0.572} & \textbf{0.416}
        & \textbf{0.353} & \textbf{0.390} & \textbf{0.172}
        & \textbf{0.344} & \textbf{0.375} \\
        \midrule
        \multicolumn{9}{c}{\textit{Qwen2.5-7B-Instruct}} \\
        \midrule
        \methodname \ Iter 1
        & 0.382 & \textbf{0.616} & 0.404 & 0.376
        & 0.345 & 0.165 & \textbf{0.432} & 0.389 \\
        \methodname \ Iter 2
        & \textbf{0.410} & 0.608 & 0.429 & \textbf{0.392}
        & 0.376 & \textbf{0.169} & 0.392 & 0.397 \\
        \methodname \ Iter 3
        & 0.397 & 0.610 & \textbf{0.438} & 0.387
        & \textbf{0.389} & 0.164 & 0.424 & \textbf{0.401} \\
        \bottomrule
    \end{tabular}
    }
    \label{tab:iters}
\end{table*}

Table~\ref{tab:iters} traces \methodname{} across three self-evolution iterations. Average EM increases monotonically from $0.354$ to $0.362$ and $0.375$ for the 3B backbone, and from $0.389$ to $0.397$ and $0.401$ for the 7B backbone. From the first to the third iteration, the largest 3B gains occur on MuSiQue ($+0.034$), 2Wiki ($+0.026$), and Bamboogle ($+0.024$), while HotpotQA improves by $0.017$. For the 7B model, 2Wiki shows the largest gain ($+0.044$), followed by PopQA ($+0.034$) and HotpotQA ($+0.011$). Individual benchmarks do not necessarily improve monotonically: for example, the 7B MuSiQue and Bamboogle scores change by $-0.001$ and $-0.008$ from the first to the third iteration, and several 7B benchmarks attain their best performance before the final iteration. Overall, repeated self-evolution consistently improves aggregate performance, although its effects vary across individual benchmarks and model scales.

\subsection{Knowledge-Graph Chain Sampler Details}
\label{app:sampler}

\begin{figure}[h]
    \centering
    \includegraphics[width=\columnwidth]{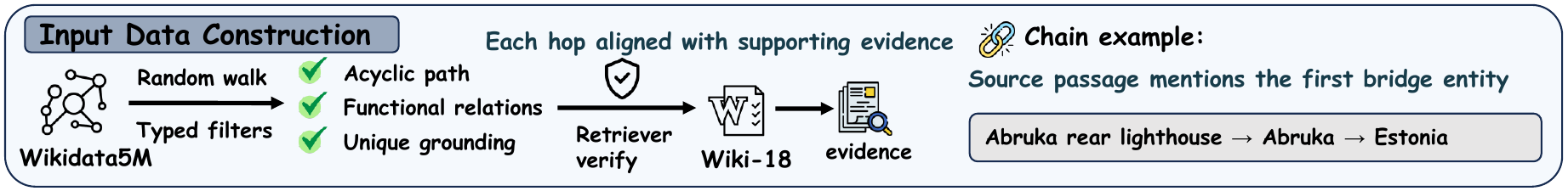}
    \caption{Knowledge-graph chain construction pipeline. Random walks on Wikidata5M generate candidate chains, which are filtered for acyclicity, functional or near-functional relations, and unique entity grounding. Retrieval verification in Wiki-18 checks that the source passage mentions the first bridge entity and that subsequent entities are reachable through per-hop retrieval.}
    \label{fig:data_construction}
\end{figure}

This appendix provides the full technical specification of the relational chain construction pipeline. The pipeline samples typed relational paths from Wikidata5M and grounds every entity in the Wiki-18 corpus, ensuring that each chain is both semantically well-typed and executable by the retriever that the solver uses.

\paragraph{Data sources.}
We use two resources. Wikidata5M~\citep{wang2021kepler} provides the relational structure: approximately $20.6$M triples over $4.8$M entities and roughly $800$ relation types, with each entity aligned to a Wikipedia article and annotated with a label and a set of aliases. The Wiki-18 corpus~\citep{karpukhin2020dense}, which contains about $21$M passages, serves two purposes: it is the retrieval index that the solver queries, and it provides a mapping from article titles to passage identifiers. The relational edges are drawn from Wikidata5M; Wiki-18 contributes only the title-to-passage mapping and the retrieval index, and it is not used as an independent source of graph structure.

\paragraph{Entity grounding and normalization.}
We build a lookup from surface strings to Wikidata entity identifiers by normalizing every label and alias in Wikidata5M: we strip parenthetical disambiguation markers (for example, converting ``Paris (city)'' to ``Paris''), apply Unicode NFKD normalization, and lowercase the result. An entity is considered groundable if its aligned article title is present in the title-to-passage mapping of Wiki-18. Entities whose normalized alias maps to more than one Wikidata identifier are marked as ambiguous and excluded during chain construction, which removes chains such as ``Paris $\rightarrow$ France'', where ``Paris'' could refer either to the city or to a mythological figure.

\paragraph{Subgraph induction and random walks.}
We first restrict Wikidata5M to the subgraph induced by groundable entities, retaining only triples whose head and tail both have articles in Wiki-18. On this induced subgraph, we perform random walks to sample chains of one to three hops. A walk is rejected if it revisits an entity (thereby forming a cycle), if it uses two or more geographic-nesting relations consecutively, or if any entity along the path lacks a retrievable source passage.

\paragraph{Typed-relation filtering.}
Wikidata5M contains many relations that are one-to-many or that otherwise do not determine a unique answer. We retain only a whitelist of functional or near-functional relation types, so that each hop maps a head entity to a well-defined target. The whitelist contains approximately $40$ relation types, organized by category in Table~\ref{tab:whitelist}. Each candidate chain is assigned a verdict. A chain is retained only if every relation belongs to the whitelist and every entity is unambiguously grounded; chains labeled as having no grounded identifier, ambiguous grounding, no typed relation, or excessive geographic nesting are discarded.

\begin{table}[t]
\centering
\caption{Representative functional and near-functional relation types retained in the whitelist for chain sampling. The whitelist contains approximately 40 relation types across five categories.}
\vspace{0.2cm}
\small
\begin{tabular}{ll}
\toprule
\textbf{Category} & \textbf{Representative Relations} \\
\midrule
Authorship \& creation & author (P50), creator (P170), composer (P86), director (P57), \\
 & screenwriter (P58), performer (P175) \\
\midrule
Geographic \& institutional & country (P17), headquarters location (P159), \\
 & country of origin (P495), location of formation (P740) \\
\midrule
Affiliations \& roles & occupation (P106), \\
 & educated at (P69), employer (P108), position held (P39) \\
\midrule
Production \& distribution & manufacturer (P176), record label (P264), \\
 & publisher (P123), production company (P272) \\
\midrule
Works \& achievements & notable work (P800), award received (P166), \\
 & field of work (P101), genre (P136) \\
\bottomrule
\end{tabular}
\label{tab:whitelist}
\end{table}

\paragraph{Two-phase retrieval verification.}
A well-typed chain is useful only if the solver can traverse it through retrieval. We verify each chain in two phases, using the same retriever and corpus that are used during training and evaluation.

\textit{Phase 1: Source-passage grounding.} For each chain, we scan the passages associated with the source entity $e_0$ (retrieved using its article title) and require that at least one passage mentions, as a word-boundary substring, the entity $e_1$ that bridges to the second hop. This guarantees that a solver starting from the source passage can observe the first bridging entity in context.

\textit{Phase 2: Per-hop retrieval probes.} For every intermediate hop $k \in \{1, \ldots, h-1\}$, we issue a retrieval query using the bridging entity $e_k$ and examine the top-$3$ retrieved passages. We require that at least one of these passages belongs to the article of the expected target entity $e_{k+1}$. This ensures that the retrieval path the solver follows can reach every entity in the chain.

A chain is retained only if the source-passage check succeeds and all per-hop retrieval probes succeed. This two-phase verification ensures that every hop of the chain is reachable through the retrieval operations that the solver actually performs.

\paragraph{Pool statistics and training usage.}
Running the construction pipeline across multiple sampling batches yields approximately $100{,}000$ verified chains. For computational and storage efficiency, we retain a fixed subset of 76,000 verified chains for the experiments reported in this paper. This subset is sampled once using a fixed random seed and shared across all methods and ablations. It is substantially larger than the number of chains consumed during training, so the additional constructed chains are not required for the reported results. During training, we sample inputs from this pool on the fly, mixing one-, two-, and three-hop chains uniformly at a $1{:}1{:}1$ ratio. The proposer is trained for $50$ update steps with a batch size of $256$, so it consumes roughly $12{,}800$ chains per iteration, which is well below the size of the retained subset. The retained pool therefore provides ample headroom, and the uniform hop sampling ensures that each batch contains a balanced mix of hop counts (with an empirical standard deviation of about $3\%$ per hop). Each sampled chain is rendered into a single or multi-turn prompt that presents the source passage together with the relational chain and its aligned evidence passages, from which the proposer generates a question and its target answer. The proposer objective and the information-gain reward are described in Sections~\ref{sec:ig_reward} and~\ref{sec:optim}.

\subsection{Format Score and Validity Criteria}
\label{app:reward}

This appendix specifies the format score $s_{\text{fmt}}$, the grounding check, and the gate that jointly control when the information-gain reward is applied.

\paragraph{Parsing.} Each proposer response is expected to contain a reasoning block followed by a question and its answer. From a response we extract an indicator $\text{has\_think}$ of whether a \texttt{<think>...</think>} block is present, together with the final \texttt{<question>} and \texttt{<answer>} spans, denoted $q$ and $a$.

\paragraph{Integrity gate.} A well-formed pair must satisfy
\begin{equation}
    \text{integrity} = \mathbb{I}\big[\, |q| > 0 \;\wedge\; |a| > 0 \;\wedge\; \operatorname{norm}(a) \not\subseteq \operatorname{norm}(q) \,\big],
\end{equation}
where $\operatorname{norm}(\cdot)$ lowercases the text and removes punctuation and articles. This requires non-empty spans and forbids the answer from appearing inside the question. A pair failing integrity receives $s_{\text{fmt}} = 0$.

\paragraph{Score aggregation.} For a pair passing the integrity gate, the score averages a base term, the presence of a reasoning block, and answer correctness:
\begin{equation}
    s_{\text{fmt}} =
    \begin{cases}
        0, & \text{integrity} = 0, \\[4pt]
        \dfrac{1 + \text{has\_think} + \text{ans\_correct}}{3}, & \text{integrity} = 1,
    \end{cases}
\end{equation}
where $\text{ans\_correct} = \mathbb{I}[\operatorname{norm}(a) = \operatorname{norm}(y)]$ compares the generated answer against the target answer $y$ fixed by the chain. The score thus takes values in $\{0, \tfrac{1}{3}, \tfrac{2}{3}, 1\}$.

\paragraph{Grounding check.} Beyond structural correctness, we apply a grounding check that verifies where the answer can be found. For a single-hop question, the answer must appear in either the source passage or one of the evidence passages. For a multi-hop question, the answer must appear in a searched evidence passage along the chain, and a pair whose answer is supported only by the seed passage is rejected. This check discourages the proposer from posing nominally multi-hop questions whose answer is already available in the source.

\paragraph{Information-gain gate and total reward.} The information-gain reward is granted only when the pair attains the maximum format score and passes the grounding check. Operationally, the IG term is retained only when $s_{\text{fmt}} = 1$ (enforced by a threshold $\gamma = 0.7$ on the discrete score) and the grounding check succeeds; otherwise it is set to zero. The full proposer reward is
\begin{equation}
    r = \lambda\, s_{\text{fmt}} + \mathbb{I}\!\left[s_{\text{fmt}} = 1 \wedge \text{grounded}\right] \cdot s_{\text{IG}},
\end{equation}
with format weight $\lambda = 0.2$, so that the format signal contributes at most $0.2$ and the information-gain reward dominates once a pair is valid. This composition ensures that the full-versus-shortcut likelihood comparison is applied only to structurally valid, correctly answered, and grounded question--answer pairs.

\subsection{Numerical Cases for Shortcut Ablation}
\label{app:shortcut-cases}

\begin{table*}[t]
    \centering

        \caption{Representative cases showing how the evaluated shortcut contexts
    prevent false-positive information-gain rewards. Here, $\Delta_{-b}$
    denotes the gain computed after excluding shortcut context $b$, whereas
    $\Delta_B$ denotes the gain computed using the complete shortcut set
    defined in ~\eqref{eq:multihop-shortcut-set}. Adding the source-only or
    individual-passage comparison changes an otherwise positive gain into a
    negative one, revealing that the target answer can already be supported
    by the corresponding shortcut context.}
    \vspace{0.2cm}
    \small
    \setlength{\tabcolsep}{6pt}
    \renewcommand{\arraystretch}{1.10}
    \resizebox{\linewidth}{!}{
    \begin{tabular}{@{}lllll@{}}
        \toprule
        Case & Shortcut context & Gain without context
        & Gain with full set $B$ & Effect \\
        \midrule

        Lake Ontario $\rightarrow$ Saint Lawrence
        & Source-only
        & $\Delta_{-\mathrm{src}}=+1.422$
        & $\Delta_B=-0.953$
        & False positive $\rightarrow$ reject \\

        Hypatima $\rightarrow$ Macquarie River
        & Individual passage
        & $\Delta_{-\mathrm{hop}}=+9.754$
        & $\Delta_B=-0.117$
        & False positive $\rightarrow$ reject \\

        \bottomrule
    \end{tabular}
    }
    \label{tab:shortcut-cases}
\end{table*}

Table~\ref{tab:shortcut-cases} provides representative numerical cases showing
how the evaluated shortcut comparisons prevent false-positive rewards. For the
Lake Ontario case, excluding the source-only context gives a positive gain of
$+1.422$, whereas including it changes the gain to $-0.953$. This result
indicates that the seed passage already provides sufficient support for the
target answer, so the question should not receive a positive information-gain
reward. Similarly, the Hypatima case has a gain of $+9.754$ when the individual
per-hop contexts are excluded, but the gain decreases to $-0.117$ after these
contexts are included. This change reveals that a single evidence passage
already provides sufficient answer information. Together, these cases
illustrate how the source-only and per-hop comparisons suppress two common
forms of shortcut: reliance on the seed passage and reliance on an individual
evidence passage.

\subsection{Representative Failure Cases of the Dr.~Zero Proposer}
\label{app:drzero-cases}
\begin{table*}[t]
  \caption{Representative Dr.~Zero proposer failures. Despite their nominal
  multi-hop labels, all three questions are generated without search and can
  be answered directly from the seed passage. Questions are reproduced
  verbatim from the generated data.}
  \vspace{0.2cm}
  \centering
  \small
  \setlength{\tabcolsep}{4pt}
  \renewcommand{\arraystretch}{1.15}
  \begin{tabularx}{\textwidth}{@{}c c c >{\raggedright\arraybackslash}X
                               >{\raggedright\arraybackslash}p{0.12\textwidth}
                               >{\raggedright\arraybackslash}X@{}}
    \toprule
    Case & Hop label & Searches & Generated question & Answer
    & Evidence in seed passage \\
    \midrule

    A1 & 3 & 0 &
    What is the fundamental group of a topological group G? &
    Abelian &
    ``The fundamental group of a topological group G is abelian.'' \\

    A2 & 2 & 0 &
    At what speed range can the Arriflex D-20 camera operate in RAW format? &
    1 to 60 frame/s &
    ``The camera is capable of running at speeds from 1 to 60 frame/s.'' \\

    A3 & 4 & 0 &
    On which beach did the 2/17th Infantry Battalion first land during the
    Allied invasion of Lae? &
    Red Beach &
    ``The 2/17th Infantry Battalion came ashore on Red Beach behind the
    2/15th.'' \\

    \bottomrule
  \end{tabularx}
  \label{tab:drzero-cases}
\end{table*}

Table~\ref{tab:drzero-cases} presents representative traces underlying the
aggregate diagnosis in Section~\ref{sec:analysis}. The generated questions
are reproduced verbatim from the proposer outputs, and the evidence column
quotes the corresponding seed passages. Although these examples are labeled
as requiring two to four hops, the proposer issues no search in any case, and
each answer is stated directly in the seed passage. The examples therefore
illustrate the semantic nature of the measured shortcut: the declared hop
count reflects the requested generation format rather than the evidence
actually required to answer the question.

\begin{table*}[t]
\centering
\caption{Representative unedited proposer generations and their corresponding construction chains. The chains are included for exposition and are not part of the generated outputs.}
\vspace{0.2cm}
\small
\setlength{\tabcolsep}{5pt}
\renewcommand{\arraystretch}{1.18}
\begin{tabularx}{0.97\textwidth}{
@{}
>{\centering\arraybackslash}p{0.05\textwidth}
>{\raggedright\arraybackslash}X
>{\raggedright\arraybackslash}p{0.29\textwidth}
>{\raggedright\arraybackslash}p{0.16\textwidth}
@{}
}
\toprule
& \textbf{Generated question}
& \textbf{Construction chain}
& \textbf{Answer} \\
\midrule
\rowcolor{gray!18}
\multicolumn{4}{@{}l@{}}{\strut\textbf{2-hop generations}} \\
\addlinespace[2pt]
(1) &
Which high-speed rail line serves the station that is a terminal for the \={O}funato Line? &
\={O}funato Line
$\rightarrow$ Ichinoseki Station
$\rightarrow$ T\={o}hoku Shinkansen &
\textbf{T\={o}hoku Shinkansen} \\
\addlinespace[2pt]
(2) &
Which family of the payload specialist that flew on STS-41-D? &
STS-41-D
$\rightarrow$ Charles D.\ Walker
$\rightarrow$ Walker family &
\textbf{Walker family} \\
\addlinespace[2pt]
(3) &
Which country is the home of the island where the Abruka rear lighthouse is located? &
Abruka rear lighthouse
$\rightarrow$ Abruka
$\rightarrow$ Estonia &
\textbf{Estonia} \\
\addlinespace[3pt]
\cmidrule(lr){1-4}
\rowcolor{gray!18}
\multicolumn{4}{@{}l@{}}{\strut\textbf{3-hop generations}} \\
\addlinespace[2pt]
(4) &
In which country is the shipwreck discussed in Hugh Edwards's book located? &
Hugh Edwards
$\rightarrow$ \emph{Islands of Angry Ghosts}
$\rightarrow$ Batavia
$\rightarrow$ Australia &
\textbf{Australia} \\
\addlinespace[2pt]
(5) &
Who is the creator of the James Bond novels that an airport from the country where Lacena Golding-Clarke represented in the Olympics is named after? &
Lacena Golding-Clarke
$\rightarrow$ Jamaica
$\rightarrow$ Ian Fleming International Airport
$\rightarrow$ Ian Fleming &
\textbf{Ian Fleming} \\
\addlinespace[2pt]
(6) &
Which fighting game marks the development company of the PlayStation game in the Mana series as its first venture into the fighting game genre? &
Mana series
$\rightarrow$ \emph{Legend of Mana}
$\rightarrow$ Square
$\rightarrow$ \emph{Tobal No.\ 1} &
\textbf{\emph{Tobal No.\ 1}} \\
\bottomrule
\end{tabularx}
\label{tab:proposer_generations}
\end{table*}

\subsection{Representative Multi-Hop Generations}
\label{app:generation_examples}
Table~\ref{tab:proposer_generations} presents qualitative examples of question--answer pairs generated by the proposer from two- and three-hop relational chains. For each example, the knowledge-graph sampler first selects a relational path and retrieves an aligned evidence passage for each entity along the path. Given the resulting chain and evidence passages, the proposer generates a question whose target answer is the terminal entity. The source entity is typically mentioned explicitly in the generated question, whereas intermediate entities remain implicit and are referred to through their relational descriptions.

For each example, we report the generated question, the target answer, and a compact representation of the underlying relational chain. The chain serves as generation metadata provided to the proposer rather than as part of the generated output, and the aligned evidence passages are omitted for readability. These examples demonstrate how the proposer verbalizes structured relational chains across different domains and hop depths, and allow direct inspection of whether the generated questions preserve the intended sequence of relations.

\subsection{Training--Test Overlap Analysis}
\label{app:data_overlap}
\begin{table}[t]
    \caption{Training--test overlap across 51{,}713 evaluation questions. A similar Q--A pair has a question Jaccard similarity of at least $0.6$ and at least one shared normalized answer alias.}
    \vspace{0.2cm}
    \centering
    \small
    \setlength{\tabcolsep}{10pt}
    \begin{tabular}{lcc}
        \toprule
        Training source
        & Similar Q--A pairs
        & Exact questions \\
        \midrule
        \methodname{}
        & 25 (0.048\%)
        & 0 (0.000\%) \\
        Search-R1
        & 1{,}108 (2.143\%)
        & 14 (0.027\%) \\
        \bottomrule
    \end{tabular}
    \label{tab:training_test_overlap}
    \vspace{-0.4cm}
\end{table}

We examine whether the performance of \methodname{} may be attributed to overlap between its generated training data and the evaluation sets. For each of the 51{,}713 test questions across the seven benchmarks, we search for training questions with a normalized token-set Jaccard similarity of at least $0.6$ and determine whether the matched questions share at least one normalized answer alias. We apply the same procedure to the 169{,}615 supervised training examples used by Search-R1 for reference.
As shown in Table~\ref{tab:training_test_overlap}, only 25 test questions (0.048\%) have both substantial lexical similarity and a shared answer with a \methodname{} training example. Manual inspection further shows that only four of these pairs ask for the same fact, and none contains an identical question. For comparison, the supervised Search-R1 training set contains 1{,}108 similar question--answer pairs and 14 exact question matches. These results indicate that direct training--test leakage is unlikely to account for the performance gains of \methodname{}.

\subsection{Robustness Across Independent Runs}
\label{sec:run-robustness}

\definecolor{dirfreerow}{RGB}{226,239,248}

\begin{table*}[t]
  \centering
  \caption{Robustness across independent training runs. Dr.~Zero results are
  taken from the single runs reported by \citet{yue2026dr}. \methodname \
  results are reported for three independent runs, followed by the mean and
  sample standard deviation. At both model scales, every \methodname \ run exceeds
  the corresponding average EM reported by Dr.~Zero.}
  \label{tab:dir-free-run-stability}
  \setlength{\tabcolsep}{3.8pt}
  \renewcommand{\arraystretch}{1.02}
  \setlength{\aboverulesep}{0.5pt}
  \setlength{\belowrulesep}{1.0pt}
  \resizebox{\textwidth}{!}{%
  \begin{tabular}{lcccccccc}
    \toprule
    Method / Run & NQ & TriviaQA & PopQA & HotpotQA & 2Wiki & MuSiQue
      & Bamboogle & Avg. \\
    \midrule

    \multicolumn{9}{c}{\textit{Qwen2.5-3B-Instruct}} \\
    \midrule
    Dr.~Zero (reported)
      & 0.397 & 0.572 & 0.431 & 0.298 & 0.291 & 0.091 & 0.200 & 0.326 \\
    \methodname \ (Run 1)
      & 0.3765 & 0.5684 & 0.4057 & 0.3545 & 0.4008 & 0.1671 & 0.3440 & 0.3739 \\
    \methodname \ (Run 2)
      & 0.3740 & 0.5777 & 0.4166 & 0.3357
      & 0.3713 & 0.1514 & 0.3200 & 0.3638 \\
    \methodname \ (Run 3)
      & 0.3795 & 0.5724 & 0.4159 & 0.3530 & 0.3903 & 0.1721 & 0.3440 & 0.3753 \\
    \rowcolor{dirfreerow}
    \textbf{\methodname \ (Mean $\boldsymbol{\pm}$ Std.)}
      & $0.3767 \pm 0.0028$
      & $0.5728 \pm 0.0047$
      & $0.4127 \pm 0.0061$
      & $0.3477 \pm 0.0104$
      & $0.3875 \pm 0.0150$
      & $0.1635 \pm 0.0108$
      & $0.3360 \pm 0.0139$
      & $\mathbf{0.3710 \pm 0.0063}$ \\
    \midrule

    \multicolumn{9}{c}{\textit{Qwen2.5-7B-Instruct}} \\
    \midrule
    Dr.~Zero (reported)
      & 0.406 & 0.608 & 0.416 & 0.362 & 0.347 & 0.104 & 0.360 & 0.372 \\
    \methodname \ (Run 1)
      & 0.3970 & 0.6104 & 0.4377 & 0.3869 & 0.3887 & 0.1638 & 0.4240 & 0.4012 \\
    \methodname \ (Run 2)
      & 0.4011 & 0.6173 & 0.4238 & 0.3730 & 0.3727 & 0.1692 & 0.4480 & 0.4007 \\
    \methodname \ (Run 3)
      & 0.3950 & 0.6212 & 0.4175 & 0.3822 & 0.3915 & 0.1746 & 0.4160 & 0.3997 \\
    \rowcolor{dirfreerow}
    \textbf{\methodname \ (Mean $\boldsymbol{\pm}$ Std.)}
      & $0.3977 \pm 0.0031$
      & $0.6163 \pm 0.0055$
      & $0.4263 \pm 0.0103$
      & $0.3807 \pm 0.0071$
      & $0.3843 \pm 0.0101$
      & $0.1692 \pm 0.0054$
      & $0.4293 \pm 0.0167$
      & $\mathbf{0.4005 \pm 0.0008}$ \\
    \bottomrule
  \end{tabular}%
  }
\end{table*}

To assess sensitivity to training randomness, we run \methodname{} with three
independent random seeds using Qwen2.5-3B-Instruct and
Qwen2.5-7B-Instruct, keeping the training budget, data-construction pipeline,
retrieval setup, and evaluation protocol fixed.
Table~\ref{tab:dir-free-run-stability} reports individual runs and their
mean and sample standard deviation. We include the single-run results reported
for Dr.~Zero~\citep{yue2026dr} as a reference; since repeated Dr.~Zero
measurements are unavailable, this comparison serves as a robustness check
rather than a formal significance test.

Across all seeds, \methodname{} exceeds the reported Dr.~Zero average EM at both
model scales. It achieves $0.3710 \pm 0.0063$ with the 3B backbone, compared
with $0.326$ for Dr.~Zero, and $0.4005 \pm 0.0008$ with the 7B backbone,
compared with $0.372$. Even the lowest-performing runs retain absolute margins
of $0.0378$ and $0.0277$, respectively. Although the 3B model exhibits greater
seed sensitivity, particularly on multi-hop benchmarks, these results show
that the improvements are not driven by a single favorable seed.

\subsection{Gold-Blind Counterfactual Judge}
\label{app:gold-blind-judge}
\begin{table}[t]
\caption{Context construction and decision rules for the gold-blind
counterfactual audit.}
\label{tab:gold_blind_judge}
\centering
\scriptsize
\setlength{\tabcolsep}{3pt}
\renewcommand{\arraystretch}{1.10}
\begin{tabularx}{\linewidth}{
@{}
>{\raggedright\arraybackslash}p{0.16\linewidth}
>{\raggedright\arraybackslash}p{0.40\linewidth}
>{\raggedright\arraybackslash}X
@{}
}
\toprule
\textbf{Condition}
& \textbf{Context and judge mode}
& \textbf{Shortcut decision rule} \\
\midrule

Closed-book
&
No evidence is supplied. The judge may answer using its parametric knowledge.
&
The judge returns \texttt{answerable=true}, and its answer matches the hidden
gold.
\\

\addlinespace[2pt]

Source-only
&
Only the seed document is supplied under the evidence-grounded prompt.
&
The judge returns \texttt{answerable=true} and \texttt{unique=true}, and its
answer matches the hidden gold.
\\

\addlinespace[2pt]

One-shot
&
The judge receives the top three passages from one retrieval using the
training retriever. Each passage is truncated to 340 whitespace-delimited
words.
&
The judge returns \texttt{answerable=true} and \texttt{unique=true}, and its
answer matches the hidden gold.
\\

\addlinespace[2pt]

Single-component
&
Each evidence-chain passage is supplied independently under the
evidence-grounded prompt.
&
A shortcut is recorded if any component independently yields a unique answer
matching the hidden gold.
\\

\bottomrule
\end{tabularx}
\end{table}

We use GPT-4o through Azure OpenAI as the gold-blind judge for the
counterfactual audit in Table~\ref{tab:shortcut_audit}. All calls use
temperature $0$ and JSON-constrained output. Each question--context condition
is evaluated in an independent call, and the judge never observes the gold
answer. For evidence-based conditions, the judge must answer solely from the
supplied context and abstain unless the answer is uniquely supported. The
closed-book condition instead permits the judge to use its parametric
knowledge. Predicted answers are compared with the hidden gold answers only
after generation. Table~\ref{tab:gold_blind_judge} summarizes the prompts,
context construction, and shortcut decision rules.

\subsection{Proposer Prompt}
\label{app:proposer_prompt}

Table~\ref{tab:complete_proposer_prompt} presents the prompt template used for proposer-side question construction. For each sampled knowledge-graph chain, the source document, relational path, and aligned evidence passages are inserted into the corresponding placeholders. These materials are available only to the proposer: the proposer generates a question--answer pair without invoking search, whereas the solver receives only the generated question and must retrieve the required evidence independently. The prompt also specifies hop-dependent composition constraints to preserve sequential dependence and prevent intermediate-entity leakage.
\begin{table*}[!h]
\caption{Summary of the proposer prompt for question construction. The proposer receives an ordered chain of entities and aligned evidence, and is instructed to generate a question without search while hiding intermediate entities and preserving sequential dependence. Positive examples, counterexamples, and detailed self-checks are omitted for brevity.}
\vspace{0.2cm}
\centering
\scriptsize
\setlength{\tabcolsep}{3pt}
\renewcommand{\arraystretch}{1.06}
\begin{tabularx}{\textwidth}{
@{}
>{\raggedright\arraybackslash}p{0.04\textwidth}
>{\raggedright\arraybackslash}X
>{\raggedright\arraybackslash}p{0.04\textwidth}
>{\raggedright\arraybackslash}X
@{}
}
\toprule
\rowcolor{gray!20}
\multicolumn{4}{@{}p{0.97\textwidth}@{}}{\strut\textbf{Core proposer instructions.} Placeholders are instantiated for each sampled knowledge-graph chain.} \\
\midrule

\multicolumn{4}{@{}p{0.97\textwidth}@{}}{\textbf{Task definition}} \\[-1pt]
\multicolumn{4}{@{}p{0.97\textwidth}@{}}{
You are an expert question writer. You are given a source document and the evidence for every hop of a pre-validated chain. All evidence is already supplied: do not search or emit \texttt{<tool\_call>}. Compose ONE question (Q) and its single unambiguous answer (A), grounded in this evidence. Output exactly one assistant turn in the form \texttt{<think>...</think><question>...</question><answer>...</answer>}.
} \\

\midrule
\rowcolor{gray!20}
\multicolumn{4}{@{}p{0.97\textwidth}@{}}{\strut\textbf{Dynamic input}} \\
\midrule

\multicolumn{4}{@{}p{0.97\textwidth}@{}}{
\begin{tabularx}{\linewidth}{
@{}
>{\raggedright\arraybackslash}X
@{\hspace{10pt}}
>{\raggedright\arraybackslash}X
@{\hspace{10pt}}
>{\raggedright\arraybackslash}X
@{}
}
\textbf{Source document (Hop 0)} &
\textbf{Ordered entity chain} &
\textbf{Per-hop evidence} \\
\texttt{\{document\}} &
\texttt{\{chain\_summary\}} &
\texttt{\{evidence\_block\}}
\end{tabularx}
\smallskip

Treat the passages as the complete construction evidence available to the proposer.
} \\

\midrule
\rowcolor{gray!20}
\multicolumn{4}{@{}p{0.97\textwidth}@{}}{\strut\textbf{Question rules}} \\
\midrule

\textbf{Q1.} &
\textbf{Real question.} Begin with a wh-word or auxiliary verb, end with exactly one question mark, and match the wh-word to the answer type. &
\textbf{Q2.} &
\textbf{No imperative.} Never begin with \emph{Identify}, \emph{Name}, \emph{List}, \emph{Describe}, \emph{Explain}, \emph{State}, \emph{Provide}, or \emph{Give}. \\

\addlinespace[2pt]
\textbf{Q3.} &
\textbf{One question.} Do not use compound constructions such as ``and what'' or ``and who.'' &
\textbf{Q4.} &
\textbf{Self-contained.} Do not refer to the document, source, passage, text, or article, or use expressions such as ``according to'' or ``as mentioned in.'' \\

\addlinespace[2pt]
\textbf{Q5.} &
\textbf{Length and leakage.} Follow the hop-specific length limit, use only one question mark, and do not reveal the final answer. &
\textbf{Q6.} &
\textbf{Hidden bridges.} Mention only Hop 0 by name. Hops $1,\ldots,n-1$ must remain implicit and appear only through descriptions grounded in the preceding hop. \\

\addlinespace[2pt]
\textbf{Q7.} &
\textbf{Surface form.} Follow the hop-specific form constraint given below. &
\textbf{Q8.} &
\textbf{Supported modifiers.} Ground every modifier in Hop 0 or a resolved hop; do not add unsupported details to force uniqueness. \\

\midrule
\rowcolor{gray!20}
\multicolumn{4}{@{}p{0.97\textwidth}@{}}{\strut\textbf{Chain-composition rules}} \\
\midrule

\multicolumn{4}{@{}p{0.97\textwidth}@{}}{
Let $\mathrm{Hop}\ 0 \rightarrow \mathrm{Hop}\ 1 \rightarrow \cdots \rightarrow \mathrm{Hop}\ n$, where Hop $n$ is the answer and Hops $1,\ldots,n-1$ are hidden bridges.
} \\[1pt]

\textbf{C1.} &
\textbf{Bounded depth.} Use zero subordinate clauses for one hop, at most one for two hops, and at most two for three hops. Never use three nested clauses or three consecutive ``of the X'' phrases. &
\textbf{C2.} &
\textbf{Sequential dependence.} Resolve each entity from the preceding entity. Use narrow entity types and concrete relations rather than vague connectors such as \emph{associated with}, \emph{related to}, or \emph{linked to}. \\

\addlinespace[2pt]
\textbf{C3.} &
\textbf{No attribute stacking.} Do not flatten multiple hops into parallel attributes. &
\textbf{C4.} &
\textbf{No mixed-hop clauses.} Each clause may describe at most one hop. \\

\addlinespace[2pt]
\textbf{C5.} &
\textbf{No single-hop disguise.} If Hop 0 already names a bridge, reframe the question so that this bridge is resolved through the preceding relation rather than directly retrieved from Hop 0. &
\textbf{C6.} &
\textbf{Functional resolution.} Each relation must uniquely determine the next entity in the direction traversed. Reject underdetermined endpoints and reverse traversals that admit multiple entities. \\

\midrule
\rowcolor{gray!20}
\multicolumn{4}{@{}p{0.97\textwidth}@{}}{\strut\textbf{Hop-specific form constraint}} \\
\midrule

\multicolumn{2}{@{}p{0.475\textwidth}@{\hspace{8pt}}}{
\textbf{One hop.}
Use 4--12 words. Begin with a wh-word or auxiliary verb, mention only the starting entity, and avoid subordinate constructions using \emph{of}, \emph{whose}, \emph{which}, or \emph{that}. Reject declarative fragments.
} &
\multicolumn{2}{p{0.475\textwidth}@{}}{
\textbf{Two or three hops.}
Use 8--18 words for two hops or 12--22 words for three hops. Mention Hop 0 exactly once and compose hidden bridges using possessives, prepositional phrases, or at most two subordinate clauses. Reject repetition, excessive nesting, overlength questions, declarative fragments, and bridge attributes stacked with \emph{and} or \emph{with}.
} \\

\midrule
\rowcolor{gray!20}
\multicolumn{4}{@{}p{0.97\textwidth}@{}}{\strut\textbf{Answer rules}} \\
\midrule

\multicolumn{4}{@{}p{0.97\textwidth}@{}}{
Output exactly one canonical answer equal to Hop $n$. Aim for 1--5 words; 6--8 are acceptable, and 9 or more are rejected. Omit leading articles unless part of a proper name (e.g., \emph{The Beatles}). Use a noun phrase, named entity, number, date, or yes/no answer; use a bare four-digit year or a clean date when applicable. Do not output a full sentence, terminal punctuation, alternatives, or placeholders such as \emph{Unknown}, \emph{N/A}, or \emph{Not specified}. Do not reveal the answer in the question.
} \\

\midrule
\rowcolor{gray!20}
\multicolumn{4}{@{}p{0.97\textwidth}@{}}{\strut\textbf{Output structure and self-check}} \\
\midrule

\multicolumn{4}{@{}p{0.97\textwidth}@{}}{
\begin{minipage}[t]{\linewidth}
\ttfamily
<think>\\[-2pt]
\hspace*{1em}Hop \(k\) \(\mid\) relation: ... \(\mid\) referent: ... \(\mid\) unique? yes, because ...\\[-2pt]
\hspace*{1em}...\\[-2pt]
\hspace*{1em}Composing: ... (compose from the innermost relation outward)\\[-2pt]
</think>\\[-2pt]
<question> the generated question </question>\\[-2pt]
<answer> the final answer </answer>
\end{minipage}
\smallskip

State the relation, referent, and uniqueness justification for each hop. If any hop is not unique, stop without emitting a question. The final question must reproduce the sentence derived in the \texttt{Composing:} line, with each relation reflected in its dependency structure and intermediate entity names hidden. Do not write literal \texttt{<question>} tags inside the reasoning block.
} \\

\bottomrule
\end{tabularx}
\label{tab:complete_proposer_prompt}
\end{table*}

\end{document}